\documentclass[11pt]{article}
\usepackage[margin=1in]{geometry}

\usepackage{amsmath,amssymb,amsfonts,amsthm}
\usepackage{booktabs}
\usepackage{array}
\usepackage{algorithm}
\usepackage{algpseudocode}
\usepackage{hyperref}
\usepackage{xcolor}
\usepackage{microtype}
\usepackage{graphicx}
\usepackage{enumitem}
\usepackage[numbers,sort&compress]{natbib}
\hypersetup{colorlinks=true, linkcolor=blue!50!black, citecolor=blue!50!black, urlcolor=blue!50!black}

\newcommand{\E}{\mathbb{E}}
\newcommand{\piold}{\pi_{\mathrm{old}}}
\newcommand{\pit}{\tilde{\pi}}
\newcommand{\kl}{\mathrm{KL}}
\newcommand{\clip}{\operatorname{clip}}
\newcommand{\ess}{\mathrm{ESS}}

\title{Free Everywhere, Exact on Trees:\\PPO's Dropped Correction Buys Sample Efficiency Under Aggressive Reuse}
\author{Nima H. Siboni\\
Juna.ai, Kastanienallee 32, 10435 Berlin, Germany\\
\texttt{nima@juna.ai}}
\date{}

\begin{document}

\maketitle

\begin{abstract}
    Common policy improvement methods, including TRPO, PPO, and GRPO, estimate policy improvement under the behavioral policy's state-visitation distribution rather than the improved policy's own. The substitution makes the objective estimable from the behavioral policy's rollouts but adds a bias growing with policy divergence, hence the trust region or clip, and hence no reuse of a batch far off-policy. We show that under history-injective dynamics, where each state is reached by exactly one history, the dropped state-visitation ratio equals the product of per-step policy ratios along the sampled prefix, on every trajectory and not only in expectation. The ratio is therefore restored exactly, from log-probabilities PPO already computes. Autoregressive generation and canonical-order constructive optimization are both history-injective. The exact correction pays importance-sampling variance that grows with the horizon, so we generalize it to a one-parameter family with PPO ($\alpha{=}0$) and the full correction ($\alpha{=}1$) as endpoints: a single bias--variance knob. A gradient-level analysis of the unclipped surrogate identifies two channels the correction acts through and three conditions under which it carries signal; an enumerable testbed confirms the conditions' predictions. On hard credit-assignment scheduling tasks, a short corrected warmup with aggressive early sample reuse learns faster than PPO and than the same reuse uncorrected; the marginal gain grows with task difficulty ($+0.02$ to $+0.09$ learning-curve AUC), and the early win over PPO tracks the prefix bias that reuse incurs. A correction held throughout, or applied where clipping already contains the reuse bias, is null to harmful.
\end{abstract}

\section{Introduction}
\label{sec:intro}

Policy-improvement methods build on the performance-difference identity for the expected discounted return $\eta$, written with the advantage $A_{\piold}$ of the old policy and discount $\gamma$,
\begin{equation}
    \eta(\pit) = \eta(\piold) + \E_{\tau \sim \pit}\!\left[ \sum_{t=0}^{\infty} \gamma^t A_{\piold}(s_t, a_t) \right],
    \label{eq:trpo-identity}
\end{equation}
whose expectation is over trajectories of the \emph{improved} policy $\pit$. The improved policy cannot be rolled out before it is known, so Conservative Policy Iteration (CPI)~\citep{kakade2002cpi} and its successors --- TRPO~\citep{schulman2015trpo}, PPO~\citep{schulman2017ppo}, and GRPO~\citep{shao2024grpo} --- evaluate the expectation under the \emph{old}, behavioral policy $\piold$ instead. The substitution makes the objective estimable from the rollout batch, but it is exact only when the two policies induce the same state-visitation distribution. Its error grows with their divergence, which is why every method in this family confines each update to a trust region or a clip and cannot reuse a batch far off-policy.

The dropped ratio has a standard unbiased estimate in any MDP, the product of policy ratios along the sampled prefix~\citep{precup2000eligibility}, computable from the log-probabilities PPO already has. Whenever the dynamics are \emph{history-injective}, each state reached by exactly one history, that product does not merely estimate the ratio but equals it on every trajectory. The contribution is not a new importance-sampling identity but this pointwise collapse on \emph{tree-structured} dynamics, together with an analysis of when paying the correction's variance improves learning. The exact correction carries trajectory importance-sampling (IS) variance that grows with the horizon, so we generalize it to a one-parameter family with PPO's biased, low-variance update at the other end.

Whether removing the bias is worth its added variance depends on the reward and on the training loop. For the unclipped surrogate, the correction changes the policy gradient through exactly two channels, both silent at the rollout policy under a conditionally centered advantage. Away from it, three conditions of increasing strength mark out where the correction has systematic leverage: the reward is decided late, it depends on the prefix, and --- for the endpoint of training rather than its speed --- the sampling loop can trap the policy.

We test these claims in two settings with different roles. An exactly enumerable sequence-model testbed is the instrument: it verifies the correction to floating-point precision and shows that the conditions predict where a partial correction wins and where it only adds variance. Constructive scheduling is the application: samples there are expensive, each rollout batch is reused, and the question is whether the correction lets early batches be reused harder. It does, on hard credit-assignment tasks: a short corrected warmup under aggressive early reuse learns faster than PPO and than the same reuse uncorrected, and the gain grows with task difficulty.

\paragraph{Contributions.} Two concurrent and independent proposals, MinPRO~\citep{lei2026prefix} and CTPO~\citep{ctpo}, apply the same correction to language models at $\alpha{=}1$; we share the base derivation with both, and each of the four claims below is absent from either (Section~\ref{sec:related-work}).
\begin{enumerate}[itemsep=0.15em, topsep=0.2em, leftmargin=1.4em]
\item The state-visitation ratio that CPI, TRPO, PPO, and GRPO drop is exactly recoverable, pointwise on every trajectory, from log-probabilities PPO already computes, whenever the dynamics are history-injective (Sections~\ref{sec:setting}--\ref{sec:correction}).
\item A one-parameter family, PPO at $\alpha{=}0$ and the exact correction at $\alpha{=}1$, turns the correction's bias--variance trade into a single knob (Section~\ref{sec:family}).
\item Three conditions --- the reward is decided late, it depends on the prefix, and the training loop admits a sampling trap --- predict when the correction helps; an enumerable testbed confirms them, and a pretrained $0.5$B model shows that the speed effect, but not the trap effect, survives realistic training (Sections~\ref{sec:mechanism}--\ref{sec:experiments}).
\item A short corrected warmup under aggressive early sample reuse learns faster than PPO on hard credit-assignment scheduling tasks; held throughout, or applied where the clip already contains the reuse bias, the correction is null to harmful (Section~\ref{sec:scheduling}).
\end{enumerate}

\section{Setting: tree-structured dynamics}
\label{sec:setting}

Let $h_t = (s_0, a_0, s_1, a_1, \ldots, s_t)$ denote the history up to time $t$ and $\phi(h_t) = s_t$ the state it reaches. \emph{History-injectivity}: $\phi$ is injective on histories of each length, so the current state identifies the history that produced it. In a general MDP many histories lead to the same state; a robot arm's joint configuration, reachable by many motion paths, is the typical non-injective case. The property concerns the state representation, not chance: a process with stochastic transitions is history-injective as long as the state records each realized outcome.

History-injectivity makes the reachable state graph a \emph{tree} rooted at the initial state; we call such processes \emph{tree-structured}. When the dynamics are also deterministic, $s_{t+1} = \mathcal{T}(s_t, a_t)$, the unique history is identified by its action sequence alone, the form both settings below take; the derivation uses only injectivity.

Autoregressive sequence models satisfy the condition natively: the state at step $i$ is the prefix $s_i = (x, y_{<i})$ and the action is the next token $a_i = y_i$, so distinct token sequences are distinct prefixes. This covers any autoregressive generator fine-tuned with reinforcement learning --- language, code, or protein sequences~\citep{ferruz2022protgpt2} --- and agentic generation in which the environment inserts tokens such as tool outputs, which are stochastic transitions that the context records.

Constructive combinatorial optimization satisfies it under one further convention. A policy builds a solution one element at a time~\citep{bello2016neural,kool2019attention}; a \emph{canonical construction order} fixes which slot is decided next (in scheduling, the machine that next becomes free, under a fixed tie-break). Each partial solution can then be constructed in only one way, so the process is history-injective. The policy still chooses the \emph{content} of every decision, only not its position.

\section{The exact correction and a family of surrogates}
\label{sec:method}

\subsection{Correcting the state-visitation distribution}
\label{sec:correction}

Written over states and actions, the performance-difference identity~(\ref{eq:trpo-identity}) is
\begin{equation}
    \eta(\pit) = \eta(\piold) + \sum_{t=0}^{\infty} \gamma^t\, \E_{s_t \sim d_{\pit, t},\, a_t \sim \pit}\!\left[ A_{\piold}(s_t, a_t) \right],
    \label{eq:pd-expanded}
\end{equation}
where $d_{\pi,t}(s) = P_\pi(S_t = s)$ is the time-$t$ state-visitation distribution of $\pi$ and $A_{\piold} = Q_{\piold} - V_{\piold}$ is the old policy's advantage, the only advantage function in this paper. Importance sampling applied twice, to the state draw and to the action draw, reweights the expectation onto rollouts of $\piold$, $s_t \sim d_{\piold,t}$ and $a_t \sim \piold$:
\begin{equation}
\begin{aligned}
    \eta(\pit) &= \eta(\piold) + \sum_{t=0}^{\infty} \gamma^t\, \E_{\piold}\bigl[ R^s_t\, r_t\, A_{\piold}(s_t, a_t) \bigr],\\
    R^s_t &= \frac{d_{\pit,t}(s_t)}{d_{\piold,t}(s_t)}, \qquad r_t = \frac{\pit(a_t \mid s_t)}{\piold(a_t \mid s_t)}.
\end{aligned}
    \label{eq:corrected-general}
\end{equation}
Equation~(\ref{eq:corrected-general}) holds in any MDP. The action ratio $r_t$ is the one PPO already uses. CPI, TRPO, and PPO drop the state-visitation ratio $R^s_t$, replacing it by $1$, because $d_{\pit,t}$ is not computable in general.

Under history-injectivity, $R^s_t$ equals the product of the per-step policy ratios along the sampled trajectory, and is therefore available from the same log-probabilities:
\begin{equation}
    R^s_t = \frac{d_{\pit, t}(s_t)}{d_{\piold, t}(s_t)} = \prod_{i<t} \frac{\pit(a_i \mid s_i)}{\piold(a_i \mid s_i)}.
    \label{eq:prefix-ratio-intro}
\end{equation}
We call $R^s_t$ the \emph{prefix ratio}. In general $d_{\pi,t}(s_t)$ sums over every history that reaches $s_t$, each weighted by a product of policy probabilities and transition kernels. History-injectivity removes the sum: exactly one history ends at $s_t$, the same for both policies, so numerator and denominator share every kernel factor; the kernels cancel, and what remains is~(\ref{eq:prefix-ratio-intro}) (Appendix~\ref{app:derivation}). Determinism plays no part: a ratio of identical kernels is one whether or not they are indicators.

The tree structure buys noise reduction, not correctness or computability. In any MDP the sampled-path weight $W_t = \prod_{i<t} \pit(a_i \mid s_i)/\piold(a_i \mid s_i)$ satisfies $\E_{\piold}[W_t \mid s_t] = R^s_t$, by standard per-decision importance sampling~\citep{precup2000eligibility}; the conditional statement is the identity behind marginalized importance sampling~\citep{liu2018curse,xie2019mis}. Tree structure adds the pointwise statement $W_t = R^s_t$ on every sampled path. Where histories merge, the conditional variance of $W_t$ given $s_t$ is the Rao--Blackwell gap between the sampled weight and the ratio it estimates~\citep{rowland2020cis}; injectivity closes that gap.

Substituting~(\ref{eq:prefix-ratio-intro}) into~(\ref{eq:corrected-general}) merges the two ratios into one product,
\begin{equation}
    R^s_t\, r_t = \prod_{i \le t} \frac{\pit(a_i \mid s_i)}{\piold(a_i \mid s_i)},
    \label{eq:full-corrected}
\end{equation}
of which PPO keeps only the factor $i = t$. Unmodified, (\ref{eq:full-corrected}) is unbiased but high-variance. With the per-step log-ratio $\ell_i = \log \pit(a_i \mid s_i) - \log \piold(a_i \mid s_i)$ and $\sigma^2$ its average per-step conditional variance, $\log R^s_t = \sum_{i<t} \ell_i$ has $\mathrm{Var}(\log R^s_t) \approx t\sigma^2$ (the centered terms form a martingale-difference sequence; Appendix~\ref{app:derivation}), so the variance of $R^s_t$ grows exponentially in the horizon: the classical trajectory-IS problem~\citep{precup2000eligibility,thomas2015highconfidence}, arising here in the policy update rather than in value estimation.

\subsection{A family of prefix-corrected surrogates}
\label{sec:family}

Let $L^s_t = \sum_{i<t} \ell_i = \log R^s_t$ be the \emph{prefix log-ratio}. Each surrogate replaces $R^s_t$ by a correction factor $g_t(L^s_t)$ (Table~\ref{tab:variants}), giving the corrected ratio $\tilde r_t = g_t(L^s_t)\, \exp(\ell_t)$. With $\pit = \pi_\theta$ the trainable policy, $\tilde r_t$ enters the usual clip-min objective
\begin{equation}
    \mathcal{L}^{\mathrm{PC}}(\theta) = \E_t\!\left[ \min\!\left( \tilde r_t \hat A_t,\, \clip(\tilde r_t, 1 {\pm} \epsilon)\, \hat A_t \right) \right],
    \label{eq:pcppo-objective}
\end{equation}
with $\epsilon$ the PPO clip range and $\hat A_t$ the advantage \emph{estimate} at position $t$, computed once from the rollout batch and held fixed across update epochs; we drop the hat only for the exact advantage.

\begin{table}[t]
\centering
\small
\setlength{\tabcolsep}{4pt}
\begin{tabular}{ll}
\toprule
Variant & $g_t(L^s_t)$ \\
\midrule
PPO        & $1$ \quad (biased; no added variance) \\
Full       & $\exp(L^s_t)$ \quad (unbiased; full IS variance) \\
Tempered   & $\exp(\alpha L^s_t)$, $\alpha \in [0, 1]$ \\
Truncated  & $\exp\!\bigl( \sum_{i = \max(0, t-k)}^{t-1} \ell_i \bigr)$ \\
Clipped    & $\exp(\clip(\alpha L^s_t, -c, c))$ \\
\bottomrule
\end{tabular}
\caption{Correction factors $g_t$ evaluated in this paper. Tempered interpolates PPO ($\alpha{=}0$) and full ($\alpha{=}1$); truncated corrects only the last $k$ steps; clipped caps the correction in log space.}
\label{tab:variants}
\end{table}

Each per-position expectation in~(\ref{eq:corrected-general}) is the position-$t$ marginal of a trajectory $\tau \sim \piold$, so the identity is a single expectation over rollouts, $\eta(\pit) - \eta(\piold) = \E_{\tau \sim \piold}[\sum_t \gamma^t R^s_t r_t A_{\piold}(s_t, a_t)]$, and each family member yields the single-trajectory estimate
\begin{equation}
    \hat L(\tau) \;=\; \sum_t \gamma^t\, g_t(L^s_t)\, r_t\, \hat A_t.
    \label{eq:Lhat}
\end{equation}
With exact advantages, $\E_{\piold}[\hat L]$ equals the true improvement only at $g_t = \exp(L^s_t)$; every other member trades this systematic error against the variance of the accumulated weight.

The added computation is the exclusive cumulative sum $L^s_t$ of the per-step log-ratios: one line of code, no extra forward or backward pass, and $O(T)$ elementwise operations per sample, a factor of the vocabulary size below the output layer's work (Algorithm~\ref{alg:tempered} and implementation in Appendix~\ref{app:family}). Computationally, then, the correction is free in any MDP; that is the sense of ``Free'' in the title, and tree structure adds exactness to it, not computability. The cost that remains is its variance. MinPRO~\citep{lei2026prefix} and CTPO~\citep{ctpo} both fix $\alpha = 1$ and add a variance-control layer, a running minimum and a $\sqrt{t}$-scaled clip respectively; each is one point of Table~\ref{tab:variants} with a stabilizer, which the family turns into one setting of a knob.

\section{When the correction matters}
\label{sec:mechanism}

This section asks where the correction enters the policy gradient, what reward structure makes it systematic rather than noise, and when a better per-update gradient changes what training reaches. Each question yields one checkable condition, C1 to C3, which delimit progressively stronger regimes of systematic leverage, not necessary conditions; Appendix~\ref{app:mechanism} gives the full statements. The trade is not monotone in $\alpha$: at finite drift an interior member can have lower variance than PPO and a better direction at once (Section~\ref{sec:exact-diagnostics}).

Let the \emph{score} $z_i(\theta) = \nabla_\theta \log \pi_\theta(a_i \mid s_i)$ and $S_t(\theta) = \sum_{i<t} z_i(\theta)$, the direction that makes the sampled prefix more probable. For the unclipped surrogate, let $\mathcal{L}^{g}_t = \E_{\piold}[g_t(L^s_t)\,r_t\,\hat A_t]$ be the position-$t$ term of $\E_{\piold}[\hat L]$. Subtracting PPO's gradient leaves
\begin{equation}
\begin{aligned}
    \Delta^g_t
    &:= \nabla_\theta \bigl(\mathcal{L}^{g}_t - \mathcal{L}^{\mathrm{PPO}}_t\bigr) \\
    &= \E_{\piold}\!\Bigl[
       r_t \hat A_t
       \bigl(
       \{g_t(L^s_t)-1\}\,z_t
       + g_t'(L^s_t)\,S_t
       \bigr)
       \Bigr].
\end{aligned}
\label{eq:gradient-two-channels}
\end{equation}
The first part, the \emph{reweighting channel}, is PPO's own gradient rescaled by how much the update has changed the probability of the sampled prefix. The second, the \emph{credit-routing channel}, has no counterpart in PPO: through $S_t$ it routes the position-$t$ advantage backward to the earlier actions that produced the prefix.

Conditioning on $s_t$, which by injectivity fixes the prefix and so $g_t(L^s_t)$ and $S_t$, replaces the action-dependent factors by $h_\theta(s_t) := \E_{\piold}[r_t \hat A_t z_t \mid s_t]$ and $\bar A_{\piold}^{\pi_\theta}(s_t) := \E_{\piold}[r_t \hat A_t \mid s_t]$, the old policy's advantage averaged under the updated policy's action draw:
\begin{equation}
\begin{aligned}
    \Delta^g_t
    = \E_{s_t\sim d_{\piold,t}}\!\Bigl[
       &\{g_t(L^s_t)-1\}\,h_\theta(s_t) \\
       &+ g_t'(L^s_t)\,S_t\,\bar A_{\piold}^{\pi_\theta}(s_t)
       \Bigr].
\end{aligned}
    \label{eq:gradient-conditional}
\end{equation}
At $\theta = \theta_{\mathrm{old}}$ both channels are silent under a \emph{conditionally centered} advantage estimate, $\E_{\piold}[\hat A_t \mid s_t] = 0$: the reweighting channel because $g_t(0) = 1$, the credit-routing channel because $\bar A^{\piold}_{\piold}(s_t)$ vanishes. The exact advantage is centered; sampled returns with a batch-mean baseline are not. The correction therefore acts through \emph{policy drift} across the reuse epochs of one update, and through \emph{imperfect centering}, present from the first step.

\paragraph{C1: the reward is decided late.} Each term of~(\ref{eq:gradient-conditional}) is a reward-blind \emph{drift factor}, $g_t(L^s_t)-1$ or $g_t'(L^s_t)S_t$, which starts at zero and grows with position, times a \emph{content factor}, $h_\theta$ or $\bar A_{\piold}^{\pi_\theta}$, which carries the advantage. The advantage at position $t$ measures only reward the actions from $t$ onward can still influence, so a reward decided by the first action vanishes from $A_{\piold}(s_t, \cdot)$ at exactly the late positions where drift is large, even if it is paid at the end. C1 requires the reward to remain \emph{contested} at positions where drift has accumulated.

\paragraph{C2: the reward depends on the prefix.} To first order the reweighting term averages to $g_t'(0)\,\mathrm{Cov}(L^s_t, h_\theta(s_t))$, nonzero only if $h_\theta$ differs between the prefixes the correction upweights and those it downweights. A reward that leaves every prefix equally good cancels across prefixes, and the credit-routing term reduces to a residual that is zero for the full correction and first-order in drift otherwise. C2 has a one-number witness, the omitted bias itself,
\begin{equation}
    b_t
    = \E_{\piold}[(R^s_t-1)\,r_t\hat A_t]
    = \mathrm{Cov}_{\piold}(R^s_t,\,r_t\hat A_t),
    \label{eq:bias-witness}
\end{equation}
measurable from any rollout batch; Section~\ref{sec:scheduling} measures it, summed over positions, as the \emph{bias dose}.

\paragraph{C3: the loop can trap the policy.} A gradient that is better in expectation need not change where training ends: PPO may reach the same solution a few updates later. The clearest amplifier is a self-reinforcing \emph{sampling trap}: a wrong early commitment makes rewarded trajectories so rare that a finite batch, scored against a batch-mean baseline, contains no restoring gradient. Whether a task admits a trap depends jointly on the reward, batch size, baseline, regularization, optimizer, and number of reuse epochs.

Read as a nested checklist, the conditions predict: with C1 inactive, little leverage; C1 without C2, a better \emph{estimator} but unchanged training; C1 and C2 without C3, faster learning to the same endpoint; C1--C3, a changed probability that a run succeeds. In every case the recovered bias must be weighed against the variance of $R^s_t$, which grows with $\alpha$ and the horizon. The analysis is of the unclipped surrogate; wherever the min in~(\ref{eq:pcppo-objective}) selects the flat clipped branch, the gradient is zero and neither channel enters.

\section{Verifying the mechanism on an enumerable testbed}
\label{sec:experiments}

\subsection{Testbed and reward ladder}
\label{sec:setup}

The testbed is an exactly enumerable autoregressive MDP with vocabulary size $V = 4$ and horizon $T = 6$ (diagnostics, control rewards) or $T = 8$ (learning). Two policy representations run on it: a small causal transformer ($793$k parameters), trained from sampled rollouts, and a tabular softmax, which admits exact $\eta$, KL, and advantages, so the correction can be verified as an identity. The compared methods differ only in $g_t(L^s_t)$ and share the objective, advantage estimation, and optimizer (Appendix~\ref{app:testbed-protocol}).

Four rewards form a ladder that switches the conditions on one at a time. Three are controls (Appendix~\ref{app:testbed-protocol}): a step-additive early gate (C1 partial, C2 weak), a \emph{late-only} reward $R=\mathbf{1}[y_{T-1}=\mathtt{B}]$ (C1 without C2), and a sparse trajectory match (C2, C3, but too rare a target for measurable training). The principal learning task is the \emph{two-branch} reward
\begin{equation}
    R(y_{0:T-1}) =
    \begin{cases}
        1 & \text{if } y_0 = \mathtt{A},\ y_{T-1} = \mathtt{B}, \\
        1 & \text{if } y_0 = \mathtt{C},\ y_{T-1} = \mathtt{D}, \\
        0 & \text{otherwise},
    \end{cases}
    \label{eq:branched-reward}
\end{equation}
which activates all three: the reward is contested until the terminal token (C1), the correct terminal action depends on the first-token branch (C2), and a wrong early commitment can make rewarded continuations disappear from a finite batch (C3), while the untrained success rate of $1/8$ keeps outcomes measurable.

\subsection{Estimator diagnostics on fixed policy pairs}
\label{sec:exact-diagnostics}

Before any training, we score the estimators on fixed policy pairs: $\piold$ against a logit perturbation $\pi_\theta$, whose magnitude dials the drift $\bar D_{\kl}(\piold,\pi_\theta)$ from $4 \times 10^{-5}$ to $0.33$. Every quantity is computed exactly: the bias, variance, and MSE of $\hat L$~(\ref{eq:Lhat}); the \emph{direction}, the cosine between $\nabla_\theta \E_{\piold}[\hat L]$ and the exact $\nabla_\theta \eta$; and the normalized effective sample size (ESS)~\citep{kong1992note} of the correction weights (definitions in Appendix~\ref{app:diagnostics}). Table~\ref{tab:diag-summary} reports the late-only reward at the largest drift, where the contrasts are sharpest.

\begin{table}[t]
\centering
\small
\setlength{\tabcolsep}{3pt}
\begin{tabular}{lccc}
\toprule
Method & $\mathrm{MSE}[\hat L]$ & $\cos$ & ESS \\
\midrule
PPO ($\alpha = 0$)             & $0.296 \pm 0.002$ & $0.20$ & $1.00$ \\
Tempered $\alpha = 0.05$       & $0.255 \pm 0.002$ & $0.26$ & $1.00$ \\
Tempered $\alpha = 0.10$       & $0.229 \pm 0.002$ & $0.33$ & $0.98$ \\
\textbf{Tempered $\alpha = 0.25$} & $\mathbf{0.203 \pm 0.003}$ & $0.53$ & $0.90$ \\
Tempered $\alpha = 0.50$       & $0.296 \pm 0.005$ & $0.81$ & $0.70$ \\
Full ($\alpha = 1$)            & $2.90 \pm 0.32$  & $1.00$ & $0.26$ \\
Clipped $(0.25,\,1.0)$         & $0.204 \pm 0.003$ & $0.52$ & $0.90$ \\
\bottomrule
\end{tabular}
\caption{Estimator diagnostics at $\bar D_{\kl} \approx 0.33$ on the late-only reward, $25$ seeds ($\pm$~SEM): MSE of the single-trajectory estimate $\hat L$~(\ref{eq:Lhat}), gradient cosine $\cos(\nabla \eta, \nabla \E[\hat L])$, and normalized ESS. $|\mathrm{Bias}|$ is $\le 3 \times 10^{-4}$ for every method and $2 \times 10^{-17}$ for the full correction (Appendix~\ref{app:diagnostics}). Bold marks the lowest MSE.}
\label{tab:diag-summary}
\end{table}

The full correction is exactly unbiased, at a steep variance price: its bias is at floating-point zero ($2 \times 10^{-17}$) at every tested drift, verifying~(\ref{eq:full-corrected}) as an algebraic identity, but its variance is an order of magnitude above PPO's and its normalized ESS falls to about a quarter. PPO's error is directional: negligible bias and the lowest variance, yet its update has cosine $0.20$ with the exact gradient, a failure invisible to bias- and variance-style diagnostics that does not average out with more data. The interior Pareto-dominates: tempered $\alpha = 0.25$ improves on PPO in direction and in MSE at once, a $32\%$ reduction at the interior minimum of the trade-off (Figure~\ref{fig:diag-mse-alpha} in Appendix~\ref{app:diagnostics}), because $\exp(0.25\, L^s_{T-1})$ partially cancels the outlier contributions PPO's last-step ratio makes at rarely visited prefixes.

\subsection{Training on the two-branch reward}
\label{sec:training}

The per-update estimator advantage compounds across iterations, and the methods order as the diagnostics predict (Table~\ref{tab:results-summary}). On the tabular policy, normalized stochastic gradient ascent (N-SGD) fixes the update magnitude, isolating direction: tempered $\alpha = 0.1$ solves the task by iteration $50$ and PPO by $70$, a $1.4\times$ difference, while the full and long-truncated corrections trail. On the transformer the outcomes are bimodal, so the advantage appears as success rate: tempered $\alpha \in \{0.05, 0.1\}$ succeeds on every seed, PPO on $7/10$. Each failed seed commits to a wrong first branch within the first few steps and never recovers --- the sampling trap of C3 (Figure~\ref{fig:h-main} in Appendix~\ref{app:h}). Adam attenuates the effect without erasing it and shifts the best variant toward clipping; at the aggressive $\mathtt{lr} = 10^{-2}$ the full correction collapses to no successful seed (Appendix~\ref{app:i}).

\begin{table}[t]
\centering
\small
\setlength{\tabcolsep}{3pt}
\begin{tabular}{lccc}
\toprule
 & Tabular & \multicolumn{2}{c}{Transformer} \\
\cmidrule(lr){2-2}\cmidrule(lr){3-4}
Method & N-SGD & N-SGD & Adam \\
\midrule
Tempered $\alpha=0.05$ & $55$ & $\mathbf{10/10}$ & $6/10$ \\
Tempered $\alpha=0.10$ & $50$ & $\mathbf{10/10}$ & $\mathbf{8/10}$ \\
Tempered $\alpha=0.25$ & $60$ & $8/10$ & $7/10$ \\
Clipped $(0.10,1.0)$ & $50$ & $9/10$ & $\mathbf{8/9}$ \\
Clipped $(0.25,1.0)$ & $50$ & $5/10$ & $7/9$ \\
PPO ($\alpha=0$) & $70$ & $7/10$ & $7/10$ \\
Full ($\alpha=1$) & $85$ & $6/10$ & $7/10$ \\
Truncated $k=4$ & $100$ & $5/10$ & $3/9$ \\
\midrule
Best $-$ PPO & $1.4\times$ & $+30\,\text{pp}$ & $+19\,\text{pp}$ \\
\bottomrule
\end{tabular}
\caption{Learning on the two-branch reward ($K=4$, $256$-trajectory rollouts, $200$ iterations). Tabular (lr $0.3$, $3$ seeds): iteration at which mean return reaches $\eta=1$. Transformer (N-SGD lr $1.0$; Adam lr $3\times10^{-3}$): fraction of seeds with final $\eta>0.5$. Bold marks the best entry per column.}
\label{tab:results-summary}
\end{table}

\subsection{Transfer to a pretrained language model}
\label{sec:pretrained-experiments}

The speed effect transfers to a pretrained policy under Adam; the trap effect does not. We fine-tune Qwen2.5-0.5B~\citep{qwen2024qwen25} on a rule-verifiable six-slot structured-record task whose rewards mirror the ladder (Appendix~\ref{app:pretrained}). On the prefix-dependent two-branch reward, tempered $\alpha = 0.10$ leads PPO by $+6.7$~pp in mean return over iterations $2$--$8$, the pre-registered endpoint, with all $16$ paired seeds favoring the correction; the gain is interior in $\alpha$ and collapses at $\alpha = 1$ (Table~\ref{tab:pretrained-summary}). On the matched late-only falsifier, which keeps the delayed reward but removes prefix dependence, the advantage contracts to $+1.9$~pp with per-seed signs at chance and the full correction is harmful on every seed. A trap cell that primes the policy onto the wrong branch does not trap it: all $16$ PPO seeds succeed, because about $8.5$ rewarded trajectories per $256$-rollout batch survive the priming. At realistic batch sizes the correction changes learning speed rather than rescuing lost runs.

\begin{table}[t]
\centering
\small
\setlength{\tabcolsep}{3pt}
\begin{tabular}{lcccc}
\toprule
 & \multicolumn{2}{c}{Two-branch} & \multicolumn{2}{c}{Late-only} \\
\cmidrule(lr){2-3}\cmidrule(lr){4-5}
Method & $\Delta$ (pp) & seeds & $\Delta$ (pp) & seeds \\
\midrule
Tempered $\alpha = 0.05$      & $+4.5$          & $9/10$           & $+1.2$ & $5/10$ \\
Tempered $\alpha = 0.10$      & $\mathbf{+6.7}$ & $\mathbf{16/16}$ & $+1.9$ & $8/16$ \\
Tempered $\alpha = 0.25$      & $+5.8$          & $10/10$          & $+1.1$ & $4/10$ \\
Clipped $(0.10,\,1.0)$        & $+6.5$          & $10/10$          & $+2.6$ & $6/10$ \\
Full ($\alpha = 1$)           & $+0.1$          & $6/10$           & $-4.4$ & $0/10$ \\
\bottomrule
\end{tabular}
\caption{Pretrained-LM study (Qwen2.5-0.5B, Adam): paired difference against PPO in mean Monte Carlo return over iterations $2$--$8$, and the paired seeds favoring the correction, on the two-branch reward and the late-only falsifier. Confidence intervals and the trap cell are in Appendix~\ref{app:pretrained}.}
\label{tab:pretrained-summary}
\end{table}

\begin{figure*}[t]
\centering
\includegraphics[width=\textwidth]{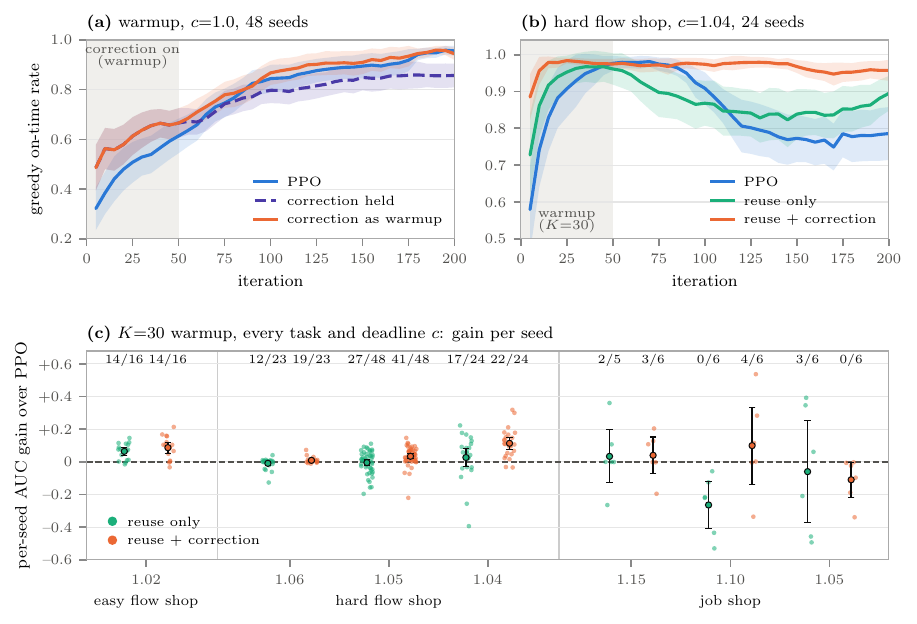}
\caption{The correction as a short warmup. \textbf{(a)} Bottleneck flow shop at the standard reuse count, $c = 1.0$: greedy validation on-time rate, mean and $95\%$ band over $48$ seeds; shaded: the $50$ warmup iterations. The warmup run leads PPO through the early climb and ends at the same level; the correction held for the whole run falls behind. \textbf{(b)} Seven-machine flow shop at its hardest deadline, aggressive early reuse ($K = 30$ for the first $50$ iterations) with and without the correction, $24$ seeds: PPO and reuse alone peak and then degrade, while the corrected warm start holds its level. \textbf{(c)} The same $K = 30$ warmup on every task and deadline: paired per-seed AUC gain over PPO, without (reuse only) and with the correction; markers are means with $95\%$ CIs, and the numbers count the seeds that beat PPO. Values, sign tests, and the correction's marginal over reuse only are in Tables~\ref{tab:sched-warmup} (a) and~\ref{tab:sched-win} (b, c), Appendix~\ref{app:sched-win}.}
\label{fig:main}
\end{figure*}

\section{Application: constructive scheduling under aggressive sample reuse}
\label{sec:scheduling}

\subsection{Setting, and the null at realistic operating points}
\label{sec:sched-null}

Scheduling is the least controlled setting in this paper and the one where samples are expensive. The expense is met by reusing each rollout batch for $K$ update epochs, and reuse moves the policy away from the data it trains on, the staleness the correction exists to repair. The tasks are flexible job-shop (FJSP), permutation flow-shop, and disjunctive job-shop instances under the canonical construction order of Section~\ref{sec:setting}, so the correction is exact. A heterogeneous graph network (HGNN) dispatcher~\citep{song2022flexible} or a transformer pointer policy constructs the schedule over $T = 12$ to $50$ decisions; the reward is a binary on-time indicator $\mathbf{1}[C_{\max} \le c \cdot C^*]$ for a deadline factor $c$ on the optimal makespan $C^*$, or a dense makespan objective (Appendix~\ref{app:scheduling}).

We compare the area under the learning curve (AUC) of the greedy validation on-time rate; $\Delta$AUC is the difference between two arms, paired by seed. Sign tests are exact binomial tests on the non-tied paired seeds, one-sided for pre-specified directional hypotheses and two-sided for falsifiers and nulls. Drift is measured after one full $K$-epoch update as $\sigma^2 T$.

At realistic operating points a held correction is null to harmful, for two measurable reasons (Table~\ref{tab:sched-null} in Appendix~\ref{app:sched-driftdial}). Realistic training accumulates no staleness to repair: at published FJSP hyperparameters the drift is $\sigma^2 T \approx 3 \times 10^{-5}$ and the correction wins $4$ of $60$ paired comparisons. And where the reward retains signal, PPO absorbs the staleness on its own: dialing the drift up five orders of magnitude helps only at a deadline where training sits at the reward floor, and a rich-reward benchmark built for the correction is null at the largest drift we measured.

\subsection{The win: a corrected warmup under aggressive early reuse}
\label{sec:sched-win}

The correction pays when it is used briefly, while the policy is still moving, and then switched off. On a bottleneck flow shop with the transformer policy, applying the correction for the first $W \in \{25, 50\}$ iterations and then running plain PPO beats PPO on AUC, while holding it for the whole run loses (Figure~\ref{fig:main}a; Table~\ref{tab:sched-warmup} in Appendix~\ref{app:sched-win}).

We raise the reuse count to $K = 30$ during the warmup because reuse-induced drift is one of the two routes through which the correction acts (Section~\ref{sec:mechanism}). Raising $K$ also changes the optimizer work per batch and how often clipping engages, so the comparison arm is the same high-$K$ warmup without the correction (\emph{reuse only}), and the correction's \emph{marginal} $\Delta$AUC, corrected minus reuse only, is what the correction contributes beyond the reuse itself; the two arms do the same optimizer work.

Aggressive early reuse pays with the correction on hard credit-assignment tasks, and much less without it. On a seven-machine flow shop with $20$ jobs and three spread-out bottlenecks, reuse alone is null at the two looser deadlines and a small win only at the tightest, while the same reuse with the correction wins at every deadline, and the marginal $\Delta$AUC grows as the deadline tightens, to $+0.087$ (Figure~\ref{fig:main}c; Table~\ref{tab:sched-win}). At the two hardest deadlines the corrected warm start also ends higher: PPO peaks and then degrades to a final on-time rate of $0.89$ and $0.79$, while the corrected run holds at $0.96$ (Figure~\ref{fig:main}b). This endpoint comparison is post hoc, and reuse without the correction also ends above PPO, so at these seed counts the higher endpoint is not separated from the reuse itself. On the easy single-bottleneck flow shop the reuse-only arm captures most of the gain and the marginal is not significant: at clip $0.2$ the clip alone contains the reuse bias, and sweeping a held $K$ from $4$ to $128$ never favors the correction (Figure~\ref{fig:ktol} in Appendix~\ref{app:sched-win}). The lever is credit-assignment difficulty, not $K$.

The largest marginal gain is on the task with the longest-range credit assignment. On a disjunctive job shop ($8$ jobs, $5$ machines, $T = 40$), at the feasible deadline the marginal $\Delta$AUC is $+0.364$ on all $6$ seeds (Figure~\ref{fig:main}c; Table~\ref{tab:sched-win}). Most of that margin is the reuse-only arm's collapse, the corrected arm's own lead over PPO is not significant, and with $n = 6$ the sign of the win is established, not its magnitude.

The early win tracks the bias dose; the late win does not. The dose is $\sum_t |b_t|$ of~(\ref{eq:bias-witness}) at the rollout policy, measured on the aggressive-reuse run without correction. Across $n = 66$ configuration--seed points with task difficulty partialled out, the first-half win over PPO rises with the dose ($r = +0.32$) while the second-half win is uncorrelated with it ($r = -0.03$; Figure~\ref{fig:wincorr} in Appendix~\ref{app:sched-win}). Across horizons $J \in \{20, \dots, 50\}$ the early lead persists while the measured prefix bias roughly doubles (Appendix~\ref{app:sched-win}).

\subsection{What kind of win, and where it stops}
\label{sec:sched-boundary}

Where PPO holds its ceiling, the warmup buys sample efficiency, not a better policy: after the off-switch the run converges to PPO's level, final performance within $0.01$, and the $\Delta$AUC is a faster early climb, about $3\%$ of AUC (Figure~\ref{fig:main}a). Where PPO itself degrades late --- the hardest flow-shop deadlines and the job shop --- the warm-started run keeps its lead through the second half of training (Figure~\ref{fig:main}b; Table~\ref{tab:sched-horizon}). Sample efficiency is not compute efficiency: the high-$K$ arms spend several times PPO's optimizer steps per batch, buying rollouts, the scarce currency here, not FLOPs. The win needs drift large enough that PPO's stale-state bias exceeds the added variance, and a reward sparse enough that PPO cannot relearn what stale data corrupted. The nulls are a property of the task and learner, not of the implementation: on an environment stripped to the testbed's structure the same code rescues trapped runs on every non-tied seed (Appendices~\ref{app:sched-controls}--\ref{app:sched-transformer}).

\section{Related Work}
\label{sec:related-work}

Two concurrent proposals apply the prefix ratio to the policy update. \citet{lei2026prefix} (MinPRO) and \citet{ctpo} (CTPO) independently name the same bias; both fix the correction at $\alpha = 1$, stabilize it with their own variance control, and evaluate on LLM mathematical-reasoning benchmarks. Neither states the structural condition under which the correction is exact, extends beyond autoregressive language models, or parameterizes the design space: the tempered family, $k$-truncation, and which point wins under which conditions are absent from both.

The product-of-ratios structure and its bias--variance trade-off are classical in off-policy evaluation~\citep{precup2000eligibility,thomas2015highconfidence}, and the standard response is to refuse the product. Marginalized importance sampling~\citep{liu2018curse,xie2019mis} estimates the state-visitation ratio instead, conditional importance sampling~\citep{rowland2020cis} frames this as Rao--Blackwellization, and \citet{liu2019opgsdc} carry the estimated ratio into the off-policy policy gradient, which is~(\ref{eq:corrected-general}) with $R^s_t$ learned rather than sampled. Retrace($\lambda$)~\citep{munos2016retrace} truncates per-step ratios in value targets; V-trace~\citep{espeholt2018impala} declines the accumulated product for the policy ratio so that ``the variance does not explode.'' Applying the sampled prefix ratio to the \emph{policy} update, where no estimate is needed, and characterizing when its variance is worth paying, is the distinguishing point.

In LLM reinforcement learning the token-level ratio is the default. GRPO~\citep{shao2024grpo} and DeepSeek-R1~\citep{deepseekr1} use it and inherit the bias we characterize; GSPO~\citep{zheng2025gspo} uses a length-normalized sequence ratio and ASPO~\citep{aspo2025} an asymmetric one; DPO~\citep{rafailov2023dpo} bypasses the ratio. None corrects the state-visitation term. Neural constructive combinatorial optimization~\citep{bello2016neural,kool2019attention}, including job-shop dispatching~\citep{song2022flexible}, trains with PPO-family learners that fix a reuse count and a clip but do not correct the state-visitation distribution of the reused batches; the canonical order that makes the correction exact is a convention these methods already use for decoding.

\section{Discussion and limitations}
\label{sec:disc}

Three nested questions decide before training whether to revisit PPO's state-visitation approximation. \emph{Is the correction exact here?} Yes under history-injective dynamics; where histories merge, as in free-order dispatching, it stays unbiased but pays a path-merging variance premium (Appendix~\ref{app:derivation}). \emph{Does it carry a systematic signal?} Only if the reward is decided late and depends on the prefix (C1, C2), which the bias witness $b_t$ measures from any rollout batch. \emph{Does the loop preserve the difference?} Only a sampling trap turns it into a success-rate effect; otherwise the correction yields at most a speed advantage (C3). In every case $\alpha$ must be matched to the horizon: long $T$, merged histories, and aggressive optimizers push the usable $\alpha$ toward zero. A heuristic operating-point argument (Appendix~\ref{app:operating-point}) suggests the effect size scales with $N/T^2$, batch size over squared horizon, which would place the scheduling nulls three orders of magnitude below the testbed's operating point; we state it as a falsifiable hypothesis, not a result.

The limitations are these. The synthetic rewards are decided at known positions, which a realistic reward model is not; the pretrained study is one $0.5$B model under one hyperparameter freeze, and the sampling trap did not transfer to it at the batch size we ran (whether it fires at batch $\le 64$ is a registered follow-up). The scheduling win needs a warmup schedule and a hard, sparse-reward task, and its largest effect rests on $6$ seeds. The full correction should not replace PPO: its variance is prohibitive at long horizons, and it loses to PPO under aggressive Adam (Table~\ref{tab:i-aggressive}). The importance-sampling identity is not new~\citep{precup2000eligibility}; what is new is its pointwise form under tree-structured dynamics and the analysis of when the variance it costs is worth paying.

\section*{AI Use Statement}
In this work, we used generative AI tools (a large language model coding and writing assistant) for developing the code for the experiments, including the scripts that check the derivations by exact enumeration, and for plotting the results; for launching and monitoring the experiment runs on cloud machines; and for improving the text of the paper. We have not used generative AI tools for the research ideas, the theoretical derivations, or the interpretation of the results. We have reviewed all AI-assisted work: AI-generated code was checked against independent tests, the corrected estimator is verified to floating-point precision against exact enumeration on the tabular testbed, and the scheduling tables are regenerated by script from the stored result files rather than transcribed; AI-improved text was checked by the authors against the derivations and result files. We take responsibility for the final content of this work, including text, claims, or artifacts produced with the aid of generative AI.

\bibliographystyle{plainnat}

\clearpage
\appendix

\section{Time-indexed versus collapsed occupancies}
\label{app:rho}

The collapsed occupancy ratio $\rho_{\pit}(s)/\rho_{\piold}(s) = \sum_t \gamma^t d_{\pit, t}(s) / \sum_t \gamma^t d_{\piold, t}(s)$ is a ratio of mixtures over time and does not simplify to a product of policy ratios; correcting it with a per-token IS product would be wrong. The product form applies at a \emph{fixed} time index: $d_{\pit, t}(s_t \mid x) / d_{\piold, t}(s_t \mid x) = \prod_{i < t} \pit(y_i \mid \cdot)/\piold(y_i \mid \cdot)$. This is why the derivation starts from the time-expanded identity rather than the collapsed $\rho$-form.

\section{Derivation of the prefix factorization}
\label{app:derivation}

\paragraph{Setup.}
Condition throughout on the initial state $s_0$. Under a policy $\pi$, a length-$t$ path $(s_0, a_0, \ldots, s_t)$ has probability $\prod_{i<t} \pi(a_i \mid s_i)\, P(s_{i+1} \mid s_i, a_i)$, and the time-$t$ state-visitation distribution is its marginal:
\begin{equation}
    d_{\pi,t}(s_t) \;=\; \sum_{\substack{\text{paths}\\ \text{ending at } s_t}} \;\prod_{i<t} \pi(a_i \mid s_i)\; P(s_{i+1} \mid s_i, a_i).
    \label{eq:path-sum}
\end{equation}
History-injectivity collapses this sum of products into a single product.

\paragraph{History-injectivity removes the sum.}
If distinct histories of length $t$ reach distinct states, exactly one path ends at $s_t$ --- the one actually taken. The sum has a single term:
\[
    d_{\pi,t}(s_t) \;=\; \prod_{i<t} \pi(a_i \mid s_i)\, P(s_{i+1} \mid s_i, a_i),
\]
with $(s_i, a_i, s_{i+1})$ the states, actions, and successors along that unique path.

\paragraph{The kernels cancel in the ratio.}
The unique path into $s_t$ is a property of the dynamics, not of the policy, so it is the \emph{same} path for $\pit$ and $\piold$, with the same transition kernels along it; only the policy probabilities differ. Dividing, every kernel factor appears identically in numerator and denominator and cancels,
\[
    R^s_t \;=\; \frac{d_{\pit,t}(s_t)}{d_{\piold,t}(s_t)} \;=\; \prod_{i<t} \frac{\pit(a_i \mid s_i)}{\piold(a_i \mid s_i)},
\]
which is~(\ref{eq:prefix-ratio-intro}): the uncomputable state-visitation ratio equals a product of per-step action ratios along the realized trajectory, available from the log-probabilities PPO already computes. However complex the dynamics may be (an arbitrary simulator, a scheduling engine, a stochastic environment), they enter only through \emph{which} path reaches $s_t$, never through the weight of the ratio.

\paragraph{Determinism is not used.}
If $s_{i+1} = \mathcal{T}(s_i, a_i)$ each kernel is an indicator, $1$ on the feasible path and $0$ otherwise, and the path weight simplifies to a pure policy product; but the cancellation above holds whatever the kernels' values, so determinism adds nothing to the identity. What it adds is descriptive: the unique history into $s_t$ is then determined by the action sequence alone. A stochastic process is history-injective whenever its state records each realized outcome --- stochastic execution in scheduling with realized durations carried in the state, or environment-inserted tokens in an agentic language-model context --- and the identity then holds with the product taken over the policy's own decisions, the environment's kernels cancelling as above.

\paragraph{Remark: what history-injectivity does --- and does not --- buy.}
Write $W_t = \prod_{i<t} \pit(a_i \mid s_i)/\piold(a_i \mid s_i)$ for the sampled-path weight, and compare it with the state ratio $R^s_t = d_{\pit,t}(s_t)/d_{\piold,t}(s_t)$ it stands in for. The two are different kinds of objects: $R^s_t$ depends only on the endpoint $s_t$, while $W_t$ depends on the whole path taken to reach it. Three statements relate them, in decreasing strength:
\begin{itemize}[itemsep=0.2em, topsep=0.2em]
\item \emph{Pointwise (needs history-injectivity).} $W_t = R^s_t$ on every sampled path --- the derivation above.
\item \emph{Conditionally unbiased (any MDP).} $\E_{\piold}[W_t \mid s_t] = R^s_t$: averaged over the paths that reach $s_t$, the weight recovers the ratio.
\item \emph{Unbiased (any MDP).} $\E_{\tau \sim \piold}[W_t\, f(s_t)] = \E_{s \sim d_{\pit,t}}[f(s)]$ for any $f$: the corrected surrogate has the right expectation.
\end{itemize}
The two general statements need no structure at all, because the transition kernels are the same under both policies: along any realized path they appear identically in the numerator and denominator of the trajectory-probability ratio and cancel, leaving $P_{\pit}(\tau)/P_{\piold}(\tau) = W_t$. Summing this over all paths ending at $s_t$ gives the conditional statement; summing over all paths gives the unbiasedness statement --- standard per-decision importance sampling~\citep{precup2000eligibility}. The conditional statement is the identity behind marginalized importance sampling~\citep{liu2018curse}, in the time-indexed form of \citet{xie2019mis}. Both require only the support condition $\piold > 0$ wherever $\pit > 0$, automatic for softmax policies.

What the structural condition buys is therefore not correctness or computability but \emph{noise}. When many paths merge into the same state, the true ratio $R^s_t$ is an average over all of them, while a single sampled path supplies one random term of that average --- right on average, wrong path by path. The conditional variance $\mathrm{Var}_{\piold}(W_t \mid s_t)$ is exactly the Rao--Blackwell gap between the sampled-path weight and the ratio it estimates, the gap that conditional importance sampling~\citep{rowland2020cis} closes by estimation. History-injectivity closes it by leaving exactly one path into each state: the sample \emph{is} the average.

In practice: when injectivity fails --- e.g., dispatching $A$ then $B$ across two free machines, or $B$ then $A$, yields the same partial schedule --- the reachability graph is a DAG, and the correction is still unbiased but pays this path-merging variance premium. A canonical construction order (Section~\ref{sec:setting}) restores injectivity by making each partial solution constructible in exactly one way; since it changes the action space, corrected-vs-uncorrected comparisons must be run within that environment. This free-order (DAG) vs.\ canonical-order (tree) distinction is used in the scheduling experiments of Appendix~\ref{app:scheduling}. Stochastic execution (random processing times, breakdowns) breaks the pointwise statement only when the state omits the realized outcome, so that different draws land on the same partial schedule and paths merge; the two general statements hold either way.

\paragraph{Variance of the prefix log-ratio.} Write $\ell_i = \mu_i + m_i$ with $\mu_i = \E[\ell_i \mid s_i]$. Since $a_i$ is drawn from $\piold(\cdot \mid s_i)$ given the past, $\E[m_i \mid s_0, a_0, \ldots, s_i] = 0$, so $(m_i)$ is a martingale-difference sequence and $\mathrm{Var}\!\left(\sum_{i<t} m_i\right) = \sum_{i<t} \E[\mathrm{Var}(\ell_i \mid s_i)] = t\sigma^2$ exactly. The approximation drops $\mathrm{Var}\!\left(\sum_{i<t} \mu_i\right) + 2\,\mathrm{Cov}\!\left(\sum_{i<t} m_i, \sum_{i<t} \mu_i\right)$, the terms carrying the variation of the conditional-mean drift across states; both vanish when $\mu_i$ is constant.

\section{Family details: implementation and unevaluated members}
\label{app:family}

The added computation is the exclusive cumulative sum $L^s_t$ of the per-step log-ratios $\ell_i$; Algorithm~\ref{alg:tempered} shows the tempered member in a standard PPO loop, and Appendix~\ref{app:family} the one-line implementation in common frameworks. The sum adds no forward or backward pass: $L^s_t$ is a linear function of log-probabilities already in the computation graph, and the backward of a cumulative sum is a reversed cumulative sum. What it adds is $O(T)$ elementwise operations per sample and one float tensor of shape $(\text{batch}, T)$, each a factor $V$ (the size of the action set, the vocabulary for an LM) below the output layer's $O(TV)$ work and the logits the forward pass materializes. Computationally, then, the correction is free, in any MDP; that is the sense of ``Free'' in the title, and tree structure adds exactness to it, not computability. The cost that remains to weigh is its variance.

\begin{algorithm}[h]
\caption{Tempered Prefix-Corrected PPO}
\label{alg:tempered}
\begin{algorithmic}[1]
\Require rollout policy $\piold$, trainable policy $\pi_\theta$, value model $V_\phi$, PPO clip $\epsilon$, prefix temperature $\alpha$, KL coefficient $\beta$
\State Sample initial states $s_0 \sim p(s_0)$ and roll out trajectories $(a_0, \ldots, a_{T-1}) \sim \piold$ (for an LM: prompts $x$ and completions $y$).
\State Store $\log \piold(a_t \mid s_t)$; compute rewards, values, and advantage estimates $\hat A_t$ (held fixed across epochs).
\For{each PPO minibatch and update epoch}
  \State Compute $\log \pi_\theta(a_t \mid s_t)$; set $\ell_t = \log \pi_\theta - \log \piold$.
  \State Compute prefix log-ratios $L^s_t = \sum_{i<t} \ell_i$ via an exclusive cumulative sum.  \Comment{the only addition to PPO}
  \State Corrected log-ratio $\tilde \ell_t = \alpha L^s_t + \ell_t$; corrected ratio $\tilde r_t = \exp(\tilde \ell_t)$.  \Comment{addition}
  \State Compute clipped policy loss~(\ref{eq:pcppo-objective}); add value loss, entropy bonus, reference-KL penalty.
  \State Update $\theta$, $\phi$.
\EndFor
\State $\piold \leftarrow \pi_\theta$; repeat.
\end{algorithmic}
\end{algorithm}

Every variant acts on the prefix log-ratio $L^s_t$ rather than on the product $R^s_t$ it exponentiates to, for four reasons. The clip acts before exponentiation, so nothing under- or overflows. $L^s_t$ is one cumulative sum over log-probabilities PPO already computes. One template covers every row of Table~\ref{tab:variants}. And the variance statement $\mathrm{Var}(\log R^s_t) \approx t\sigma^2$ lives in log space.

In code, the correction is one line in any framework (\texttt{ell} of shape $(\text{batch}, T)$): PyTorch \texttt{L = torch.cumsum(ell, dim=-1) - ell}; JAX \texttt{L = jnp.cumsum(ell, axis=-1) - ell}; TensorFlow \texttt{L = tf.cumsum(ell, axis=-1, exclusive=True)}. Two further members of the family are natural but not evaluated here. A \emph{self-normalized} variant divides each weight by the batch mean of the weights --- the classical weighted importance-sampling estimator~\citep{hesterberg1995weighted,mahmood2014weighted,swaminathan2015self}; its bias is a small-batch artifact, $O(1/|B|)$, unlike PPO's structural bias, which no batch size removes. An \emph{adaptive} variant selects the largest $\alpha$ each batch subject to a floor on the effective sample size of the reweighted batch, the standard weight-degeneracy diagnostic of importance sampling~\citep{kong1992note}, used analogously to constrain policy updates by \citet{metelli2018pois}.

\section{Full statements of the conditions}
\label{app:mechanism}

This appendix gives the complete arguments behind the conditions summarized in Section~\ref{sec:mechanism}; notation as there. Throughout, \emph{prefix} means the partial trajectory up to position $t$, the history $h_t$ of Section~\ref{sec:setting}; at positions where the environment rather than the policy moves, $\ell_i$ and $z_i$ are zero, so $L^s_t$ and $S_t$ both run over the policy's own decisions.

\paragraph{Both channels at the rollout policy.} At $\theta = \theta_{\mathrm{old}}$, the first gradient step after a rollout, both channels of~(\ref{eq:gradient-conditional}) are silent under a conditionally centered advantage estimate. The reweighting channel vanishes pointwise, since $L^s_t = 0$ and $g_t(0) = 1$. The credit-routing channel is controlled by $\bar A_{\piold}^{\piold}(s_t)$, the value of $\bar A_{\piold}^{\pi_\theta}$ at $\theta_{\mathrm{old}}$ where $r_t = 1$, which reduces to $\E_{\piold}[\hat A_t \mid s_t]$; the channel vanishes state by state exactly when the advantage estimate is \emph{conditionally centered}, zero-mean given the state. The exact advantage is, by definition ($\E_{a \sim \piold}[A_{\piold}(s_t, a)] = 0$), and so is any estimator whose baseline equals the true state value, such as returns-to-go corrected by a well-fit critic. The estimators training uses are not: sampled returns with a batch-mean baseline are centered across the batch, not within each state, leaving $\E_{\piold}[\hat A_t \mid s_t] \neq 0$. The correction therefore acts through two routes: \emph{policy drift}, the movement away from the rollout policy across the reuse epochs of one update or on stale data; and \emph{imperfect centering} of the advantage estimate, which is present from the first step.

\paragraph{C1.} The first condition is that the reward must still be undecided late (C1). Each term of~(\ref{eq:gradient-conditional}) is a product of a \emph{drift factor}, $g_t(L^s_t)-1$ or $g_t'(L^s_t)\,S_t$, and a \emph{content factor}, $h_\theta$ or $\bar A_{\piold}^{\pi_\theta}$, at the same position, and the two have separate origins. The drift factors are reward-blind: they start at zero and grow with position ($L^s_0 = 0$ and $\mathrm{Var}(L^s_t) \approx t\sigma^2$ by Section~\ref{sec:correction}; $S_t$ collects $t$ scores), and depend on the reward only through earlier gradient steps. The reward enters through the content factors, which carry the advantage, and the advantage at position $t$ measures only reward that position can still influence. $A_{\piold}(s_t, a)$ is the difference between the expected return given $(s_t, a)$ and given $s_t$ alone, so any reward component whose expectation given $s_t$ no longer depends on the actions from $t$ onward contributes equally to both and cancels: reward already collected, and equally reward not yet paid that the actions ahead can no longer influence. What sets where content can live is therefore the position at which the reward is \emph{decided}, not the position at which it is delivered. A reward decided by the first action vanishes from $A_{\piold}(s_t, \cdot)$ for every $t \ge 1$, exactly the late positions where the drift factors are large, even if it is paid at the terminal step, whereas a reward that late actions can still win or lose keeps the content factors alive there.\footnote{For the exact advantage or any conditionally centered estimate. An imperfectly centered estimate, such as the trajectory return minus a batch-mean baseline, is constant along the trajectory and does survive at late positions. But an early-decided reward is fixed by the prefix, so conditioning on $s_t$ makes $\hat A_t$ a function of the state alone; this still silences the reweighting channel, $h_\theta(s_t) = \hat A(s_t)\,\E_{\piold}[r_t z_t \mid s_t] = 0$ since $\E_{\piold}[r_t z_t \mid s_t] = \nabla_\theta \sum_a \pi_\theta(a \mid s_t) = 0$, while the credit-routing content $\bar A_{\piold}^{\pi_\theta}(s_t) = \hat A(s_t)$ stays alive --- the imperfect-centering route above.} C1 is this alignment: the reward must remain \emph{contested}, still dependent on the actions ahead, at positions where drift has accumulated. In the rewards studied below the deciding and the delivering position coincide, so we keep the shorthand \emph{late reward}; the operative property is late \emph{decision}.

\paragraph{C2.} The second condition is that the reward must depend on the prefix (C2). At a fixed position, (\ref{eq:gradient-conditional}) still averages over states. To first order $g_t(L)-1 \approx g_t'(0)\,L$, so the reweighting term's average is, to leading order in the drift, $g_t'(0)\,\mathrm{Cov}(L^s_t,\,h_\theta(s_t))$. C2 is the requirement that this covariance be nonzero: $h_\theta$ must differ between the states the correction upweights ($L^s_t > 0$) and those it downweights ($L^s_t < 0$). A reward that leaves every prefix equally good makes the conditional law of $(a_t, \hat A_t)$ given $s_t$ the same at every state. Then $h_\theta$ varies across states only through the score, so the covariance carries no reward signal, and $\bar A_{\piold}^{\pi_\theta}$ is a constant $c$ across states, so the credit-routing average reduces to $c\,\E_{\piold}[g_t'(L^s_t)\,S_t]$. For the full correction this factor is exactly zero at any $\theta$: $g_t' = R^s_t$ moves the prefix draw to $\pi_\theta$, where $\E_{\pi_\theta}[z_i] = 0$. For tempered members it is zero at $\theta = \theta_{\mathrm{old}}$, where $\E_{\piold}[S_t] = 0$ by the same score identity, and first-order in the drift beyond. A conditionally centered advantage makes $c$ first-order as well, leaving a second-order residual; imperfect centering keeps $c$ order one, the centering route again. The late-only reward of Section~\ref{sec:setup} instantiates this case, C1 active and C2 absent.

\paragraph{C3.} The third condition is that the sampling-and-update loop must amplify a per-update gradient difference rather than wash it out (C3). A gradient that is better in expectation need not change where training ends: PPO may learn the same solution a few updates later, so without amplification the correction shows up as speed, not as a different endpoint. The clearest amplifier is a self-reinforcing \emph{sampling trap}: a wrong early commitment makes rewarded trajectories so rare that a finite batch, scored against a batch-mean baseline, contains no restoring gradient, so the next batch is drawn from a still-worse policy. Whether a task admits such a trap depends jointly on the reward, the batch size, the baseline, entropy or KL regularization, the optimizer, and the number of reuse epochs.

\begin{table}[t]
\centering
\small
\begin{tabular}{>{\raggedright\arraybackslash}p{0.16\linewidth}>{\raggedright\arraybackslash}p{0.28\linewidth}>{\raggedright\arraybackslash}p{0.26\linewidth}>{\raggedright\arraybackslash}p{0.19\linewidth}}
\toprule
Condition & Plain statement & Quantity to check & If it fails \\
\midrule
C1: late-decided reward & reward still action-dependent after prefix mismatch has accumulated & $|L^s_t|$ at advantage-carrying positions ($\mathrm{Var}(L^s_t) \approx t\sigma^2$) & channels are inert; only variance remains \\
C2: prefix-dependent reward & upweighted and downweighted prefixes differ in value & $b_t = \mathrm{Cov}_{\piold}(R^s_t, r_t \hat A_t) \neq 0$ & contributions cancel across prefixes \\
C3: sampling trap & an early mistake starves batches of rewarded trajectories & frequency of zero-reward batches & speed gain at most; same endpoint \\
\bottomrule
\end{tabular}
\caption{The three conditions under which the correction matters, the measurable quantity behind each, and what is lost when the condition fails. C1 and C2 concern the per-update gradient; C3 concerns the repeated sampling-and-update process.}
\label{tab:conditions}
\end{table}

In every case the recovered bias must be weighed against the variance of the prefix ratio $R^s_t$, which grows with $\alpha$ and with the horizon (Section~\ref{sec:correction}); the conditions say when there is a bias worth recovering, not that recovering it is free.

Algorithm~\ref{alg:tempered} optimizes the clipped objective~(\ref{eq:pcppo-objective}), not the surrogate analyzed above, and clipping gates both channels: wherever the min in~(\ref{eq:pcppo-objective}) selects the flat clipped branch, the position-$t$ gradient is zero and neither $g_t$ nor $g_t'$ enters. The observation of Section~\ref{sec:scheduling} that on an easy task the clip alone contains the bias of even aggressive reuse describes this gate empirically; it is not derived here.

\section{An operating-point hypothesis for the effect size}
\label{app:operating-point}

The three questions decide whether an effect is possible; a heuristic operating-point argument suggests its magnitude, and we state it as a hypothesis. Expanding the tempered estimator around $\theta_{\mathrm{old}}$ ($e^{\alpha L^s_t} \approx 1 + \alpha L^s_t$), the reweighting channel's expected contribution is $\alpha S$ with $S = \sum_t \mathrm{Cov}(L^s_t,\, h_\theta(s_t))$, the omitted bias itself, while the credit-routing channel's variance grows like $T^2$ (position $t$ sums $t$ score vectors) even at zero drift. With $N$ trajectories per batch and gradient scale $G$, the expected per-update progress advantage then has the form
\begin{equation}
    \Delta(\alpha) \;\approx\; \alpha\, S \;-\; \frac{\alpha^2}{N}\, V,
    \qquad V \;\propto\; T^2\, G\, (1 + \sigma^2),
    \label{eq:dominance}
\end{equation}
maximized at $\alpha^\star = S N / 2V$ with peak $\Delta^\star = S^2 N / 4V$. The peak advantage carries a prefactor of order $N/T^2$ if $S$ scales with $T$ more slowly than $V$ does --- writing $S^2/V$ as $\rho^2\sigma^4/T^2$ up to the unresolved $T$-dependence of $S$, with $\rho$ the correlation between the drift direction and reward-relevant structure ---
\begin{equation}
    \Delta^\star \;\propto\; \rho^2 \sigma^4 \cdot \frac{N}{T^2},
    \label{eq:peak-advantage}
\end{equation}
and (\ref{eq:peak-advantage}) is the hypothesis: the $T$-scaling of $S$ is not pinned down here. Its content is that batch size relative to squared horizon decides whether a correct-sign effect clears the seed-noise floor. The testbed cells operate at $N/T^2 \approx 256/6^2 \approx 7$; the scheduling experiments at $20/50^2 \approx 0.008$, three orders of magnitude lower at nominally comparable drift, correlation, and reward shape. Under the hypothesis, the scheduling nulls at realistic operating points are consistent with the mechanism rather than evidence against it, and the hypothesis is falsifiable: a paired advantage that grows linearly in $N$ and falls as $1/T^2$, which a short-horizon scheduling variant ($T \approx 20$) at testbed batch sizes ($N = 256$) would test.

\section{Testbed protocol and estimator diagnostics}
\label{app:testbed-protocol}

\subsection{Common protocol}

The transformer policy used in the synthetic learning experiments is a $4$-layer, $d_{\mathrm{model}}=128$, $4$-head causal transformer ($\approx 793$k parameters). The tabular policy has one logit per prefix--token pair, $V(V^T{-}1)/(V{-}1)$ in total: about $6$k at $T = 6$ and $90$k at $T = 8$. Unless stated otherwise, learning experiments use $256$ trajectories per iteration, returns-to-go advantages with a batch-mean baseline, and $K=4$ update epochs. The compared estimator variants are PPO; tempered $\alpha\in\{0.05,0.1,0.25,0.5,1\}$; truncated $k\in\{1,2,4\}$; and clipped $(\alpha,c)\in\{(0.1,1.0),(0.25,1.0),(0.25,2.0)\}$. Sign tests follow the convention of Section~\ref{sec:scheduling}: exact, one-sided for pre-specified directional improvements, two-sided for falsifier and null checks.

\paragraph{Control rewards of the ladder.} The three controls of Section~\ref{sec:setup}, each stopping short of either the full condition set or measurable training outcomes:
\begin{itemize}[itemsep=0.25em, topsep=0.25em]
\item \textbf{Step-additive early gate.} The first token gates the reward: for sequences starting with $y_0=\mathtt{A}$ (resp.\ $y_0=\mathtt{C}$) the target symbol is $\mathtt{B}$ (resp.\ $\mathtt{D}$), the reward is the fraction of later positions equal to the target, $R=\frac{1}{T-1}\sum_{t=1}^{T-1}\mathbf{1}[y_t=\text{target}]$, and any other first token gives $R=0$. Signal therefore accrues at every step, only partly overlapping with accumulated prefix mismatch (C1 partial), and prefix dependence is weak (C2 weak), making the early gate the lowest-leverage control.
\item \textbf{Late-only.} $R=\mathbf{1}[y_{T-1}=\mathtt{B}]$ is decided by the final action, the position where the prefix ratio $R^s_t$ is largest (C1), but the reward does not distinguish earlier prefixes (C2 absent): this is the rung that isolates C1 from C2.
\item \textbf{Sparse trajectory match.} $R=\mathbf{1}[y=\mathtt{ABBBBB}]$ rewards exactly one complete sequence at $T=6$, adding what the late-only rung lacks: prefix dependence (C2), and with it the possibility of a narrow sampling trap (C3). But a single rewarded sequence is too rare a target --- probability $4^{-6}$ under the uniform policy --- and training outcomes on it are noisy.
\end{itemize}

\subsection{Estimator diagnostics}
\label{app:diagnostics}

\begin{figure}[t]
\centering
\includegraphics[width=0.55\linewidth]{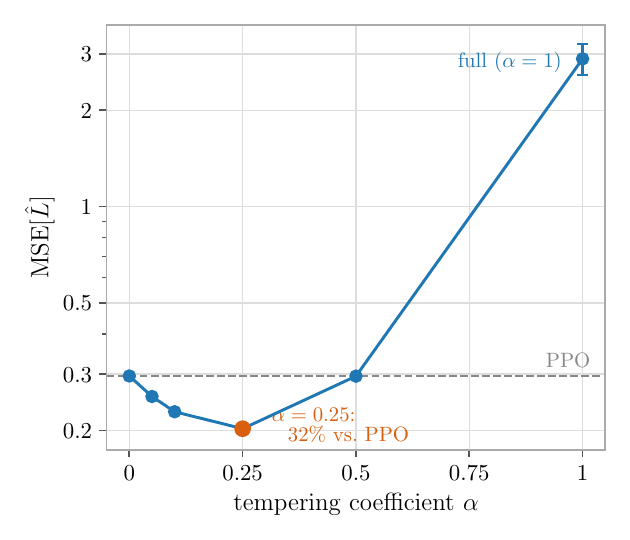}
\caption{Mean squared error $\mathrm{MSE}[\hat L]$~(\ref{eq:mse-def}) of the single-trajectory estimate versus the tempering coefficient $\alpha$, on the late-only reward at the largest tested drift ($\bar D_{\kl} \approx 0.33$, $25$ seeds, error bars $\pm 1$~SEM, log scale). The dashed line is PPO ($\alpha = 0$). The trade-off has an interior minimum at tempered $\alpha = 0.25$; the values are in Table~\ref{tab:diag-summary}.}
\label{fig:diag-mse-alpha}
\end{figure}

Each estimator is scored first as an estimator: the bias, variance, and mean squared error (MSE) of the single-trajectory estimate $\hat L$~(\ref{eq:Lhat}),
\begin{align}
    \mathrm{Bias}[\hat L] &= \E_{\piold}[\hat L] - \bigl(\eta(\pi_\theta) - \eta(\piold)\bigr), \label{eq:bias-def}\\
    \mathrm{Var}[\hat L]  &= \E_{\piold}\!\left[\bigl(\hat L - \E_{\piold}[\hat L]\bigr)^2\right], \label{eq:var-def}\\
    \mathrm{MSE}[\hat L]  &= \E_{\piold}\!\left[\bigl(\hat L - (\eta(\pi_\theta) - \eta(\piold))\bigr)^2\right] = \mathrm{Bias}[\hat L]^2 + \mathrm{Var}[\hat L], \label{eq:mse-def}
\end{align}
where averaging $\hat L$ over a batch of $N$ trajectories divides the variance by $N$ and leaves the bias unchanged. Two further diagnostics look past the scalar estimate. The \emph{direction}, the cosine $\cos\bigl(\nabla_\theta \eta,\, \nabla_\theta \E_{\piold}[\hat L]\bigr)$ between the gradient of the expected estimate and the exact return gradient, scores the update the estimator induces. The \emph{normalized effective sample size}, $\ess = \left(\E_{\piold}[w]\right)^2 / \E_{\piold}[w^2] \in (0, 1]$ with $w$ the estimator's prefix-correction weight $g_t(L^s_t)$~\citep{kong1992note}, scores the batch: $N$ trajectories carry the information of roughly $\ess \cdot N$ equally weighted ones. The contrasts are sharpest on the late-only reward at the largest tested drift; Appendix~\ref{app:diagnostics} reports the sweeps over all drift magnitudes and both control rewards, and Table~\ref{tab:diag-summary} and Figure~\ref{fig:diag-mse-alpha} the full variant grid at this operating point, $\bar D_{\kl} \approx 0.33$.

\begin{table}[t]
\centering
\small
\begin{tabular}{lcccc}
\toprule
Method & $|\mathrm{Bias}[\hat L]|$ & $\mathrm{MSE}[\hat L]$ ($\pm$~SEM) & $\cos(\nabla \eta, \nabla \E[\hat L])$ & Normalized ESS \\
\midrule
PPO ($\alpha = 0$)             & $1.8\times10^{-4}$ & $2.96\times10^{-1}\pm 1.9\times10^{-3}$ & $0.20$ & $1.00$ \\
Tempered $\alpha = 0.05$       & $2.3\times10^{-4}$ & $2.55\times10^{-1}\pm 1.9\times10^{-3}$ & $0.26$ & $1.00$ \\
Tempered $\alpha = 0.10$       & $2.6\times10^{-4}$ & $2.29\times10^{-1}\pm 2.2\times10^{-3}$ & $0.33$ & $0.98$ \\
\textbf{Tempered $\alpha = 0.25$} & $3.0\times10^{-4}$ & $\mathbf{2.03\times10^{-1}\pm 2.7\times10^{-3}}$ & $0.53$ & $0.90$ \\
Tempered $\alpha = 0.50$       & $2.6\times10^{-4}$ & $2.96\times10^{-1}\pm 5.2\times10^{-3}$ & $0.81$ & $0.70$ \\
Full ($\alpha = 1$)            & $2.1\times10^{-17}$& $2.90\times10^{0}\pm 3.2\times10^{-1}$  & $1.00$ & $0.26$ \\
Clipped $(0.25,\,1.0)$         & $2.8\times10^{-4}$ & $2.04\times10^{-1}\pm 2.6\times10^{-3}$ & $0.52$ & $0.90$ \\
\bottomrule
\end{tabular}
\caption{Estimator diagnostics at $\bar D_{\kl} \approx 0.33$ on the late-only reward, $25$ seeds; single-trajectory estimate $\hat L$~(\ref{eq:Lhat}). Bold marks the lowest MSE.}
\label{tab:diag-full}
\end{table}

\paragraph{Measurement procedure.} Each measurement is made on a controlled pair of policies standing in for the data-collecting and updated policies of one reuse step. The rollout policy $\piold$ is a fixed near-uniform tabular softmax; the candidate $\pi_\theta$ is obtained by adding Gaussian noise to every logit, $\theta_{\pit} = \theta_{\piold} + \delta\,\varepsilon$ with $\varepsilon \sim \mathcal{N}(0, I)$, so the magnitude $\delta$ dials the policy drift directly. We sweep $\delta \in \{0.01, 0.03, 0.1, 0.3, 0.6, 1.0\}$, spanning $\bar D_{\kl}(\piold, \pit)$ from $4 \times 10^{-5}$ to $0.33$, with $5$ noise seeds per magnitude ($25$ for the late-only run of (d)); all $12$ variants are evaluated on the same policy pair within each (seed, $\delta$) cell.

Because the MDP is enumerable at horizon $T=6$, every quantity is computed exactly by summing over all $4^6$ sequences, so no Monte Carlo error enters. For each variant, the single-trajectory estimate $\hat L$~(\ref{eq:Lhat}) is formed from the exact backward-induction advantages $A_{\piold}$, and its expectation is taken under the exact sequence distribution of $\piold$. Bias, variance, MSE, gradient cosine, and normalized ESS are as defined in Section~\ref{sec:exact-diagnostics}; the two gradients in the cosine are computed by automatic differentiation through the exact sums and flattened over the candidate's full logit table, one cosine per policy pair, and the normalized ESS is taken over the joint (sequence, position) distribution.

\paragraph{(a) PPO.} On the step-additive early-gate reward, PPO's bias is negative and grows monotonically with KL ($-5.3 \times 10^{-7}$ to $-4.8 \times 10^{-3}$); variance stays near the trajectory-advantage floor ($3.3 \times 10^{-2}$ to $5.4 \times 10^{-2}$); gradient cosine degrades from $1.000$ to $0.951$. Bias stays below $5 \times 10^{-3}$ --- masked by the variance floor in MSE but visible in direction.

\paragraph{(b) Full.} Bias bounded by $2 \times 10^{-17}$ at every $\delta$ --- floating-point zero, verifying~(\ref{eq:full-corrected}) as an algebraic identity, a check no sampled experiment could provide. Gradient cosine stays $1.000$. The cost: variance grows to $1.4 \times 10^{-1}$ ($2.7\times$ PPO at the same KL) and normalized ESS collapses from $\approx 1.0$ to $\approx 0.27$.

\paragraph{(c) Intermediate variants.} Interpolation is monotone in $\alpha$, $k$, and $c$ on every metric. At the largest $\delta$: interior MSE minimum at tempered $\alpha = 0.25$ / clipped $(0.25, 1.0)$ ($5.0 \times 10^{-2}$, vs.\ PPO's $5.4 \times 10^{-2}$ and full's $1.4 \times 10^{-1}$ --- a $7\%$ margin on this reward, capped by the variance floor); gradient-cosine ordering full $1.000 >$ truncated $k{=}4$ $0.998 >$ tempered $0.5$ $0.985 >$ tempered $0.25$ $0.973 >$ PPO $0.951$; nESS $0.90$ (tempered $0.25$), $0.70$ ($0.5$), $0.29$ (truncated $k{=}4$), $0.27$ (full). Gradient direction is already decisive here --- the prediction taken to the training experiments.

\paragraph{(d) Late-only reward --- sharpened gaps.} Concentrating all advantage at $t = T-1$ via $R = \mathbf{1}[y_{T-1} = \mathtt{B}]$ amplifies everything (Table~\ref{tab:diag-summary}, Figure~\ref{fig:diag-mse-alpha}). At $\bar D_{\kl} \approx 0.33$: PPO's bias stays $\lesssim 2 \times 10^{-4}$ but its gradient cosine collapses to $0.20$, near-orthogonal; full's variance explodes to $2.9$, an order of magnitude above PPO, driven by trajectory-IS outliers at rare-visit prefixes.

Tempered $\alpha = 0.25$: MSE $2.03 \times 10^{-1}$ ($32\%$ below PPO), variance $2.02 \times 10^{-1}$ against PPO's $2.96 \times 10^{-1}$ (Table~\ref{tab:diag-summary}, Figure~\ref{fig:diag-mse-alpha}). A terminal reward that late tokens still decide --- final-answer correctness, a terminal reward-model score --- falls in this regime; one already sealed by the early tokens falls under the early-decided case of Section~\ref{sec:mechanism}.

\subsection{Null controls: rewards without prefix dependence or a sampling trap}
\label{app:diagnostics-null-controls}

On the three control rewards of the ladder (Section~\ref{sec:setup}) the estimator MSE margin of the best interior variant over PPO is $11\%$ on the step-additive early gate (C1 partial, C2 weak); $32\%$ on late-only (C1 without C2), with no training-loop advantage --- the cleanest separation of the estimator diagnostic from the training effect; and present, with noisy training outcomes, on the sparse trajectory match (C2, C3). The two-branch reward~(\ref{eq:branched-reward}) activates C1--C3 while keeping rewarded trajectories frequent enough for training outcomes to be measurable, which is why the training sub-experiments use it.

\section{Testbed training experiments}
\label{app:training}

\subsection{Tabular policy under N-SGD}
\label{app:g}

Tabular policy, two-branch reward, N-SGD at $\mathtt{lr} = 0.3$, $K = 4$, $3$ seeds, $200$ iterations. N-SGD removes the direction-vs-magnitude confound.

\begin{figure}[t]
\centering
\includegraphics[width=0.7\linewidth]{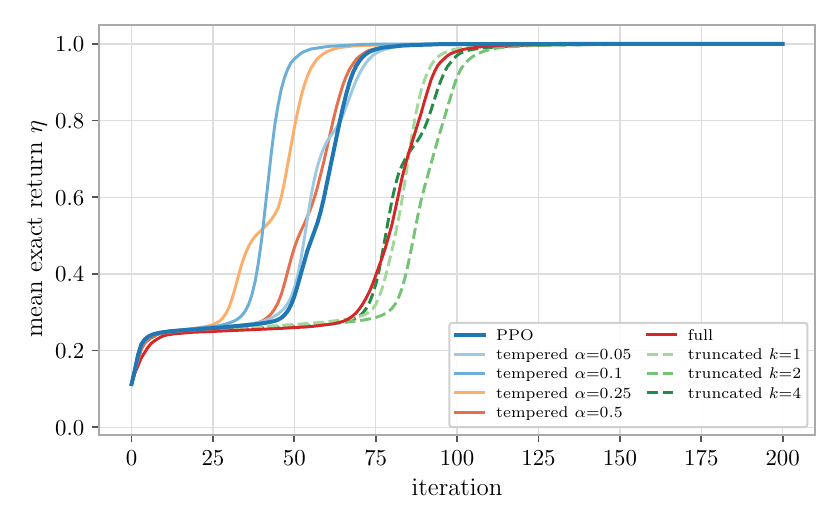}
\caption{Exact return versus iteration on the two-branch reward: tabular policy, N-SGD ($\mathtt{lr} = 0.3$, $K = 4$, mean over $3$ seeds). The clipped variants are omitted: their curves overlap exactly with those of their tempered counterparts in this run, as the log-space cap $c$ never binds.}
\label{fig:g-detail}
\end{figure}

The break-through order through the $\eta \approx 0.25$ plateau is strict and matches the estimator prediction: interior tempered and clipped first, PPO next, full and truncated $k = 4$ last, paying the predicted variance penalties (Figure~\ref{fig:g-detail}, Table~\ref{tab:g-detail}).

\begin{table}[h]
\centering
\small
\begin{tabular}{lccc}
\toprule
Method & Break-through iter.\ & $\eta$ at iter.\ $50$ & Iter.\ to $\eta = 1.000$ \\
\midrule
\textbf{Tempered $\alpha = 0.10$} & $30$--$45$ & $\mathbf{0.961}$ & $50$ \\
Clipped $(0.25,\,1.0)$            & $30$--$45$ & $0.960$          & $50$ \\
Tempered $\alpha = 0.05$          & $50$       & $0.900$          & $55$ \\
Tempered $\alpha = 0.25$          & $50$--$55$ & $0.781$          & $60$ \\
Tempered $\alpha = 0.50$          & $50$--$60$ & $0.468$          & $70$ \\
PPO                               & $50$--$60$ & $0.341$          & $70$ \\
Full                              & $70$       & $0.260$          & $85$ \\
Truncated $k = 4$                 & $80$--$90$ & $0.262$          & $100$ \\
\bottomrule
\end{tabular}
\caption{Tabular policy under N-SGD ($\mathtt{lr} = 0.3$, $K = 4$, $3$ seeds): iteration of break-through past the $\eta \approx 0.25$ plateau, exact return at iteration $50$, and iteration at which $\eta$ reaches $1$.}
\label{tab:g-detail}
\end{table}

\subsection{Transformer under N-SGD}
\label{app:h}

\begin{figure}[t]
\centering
\includegraphics[width=0.7\linewidth]{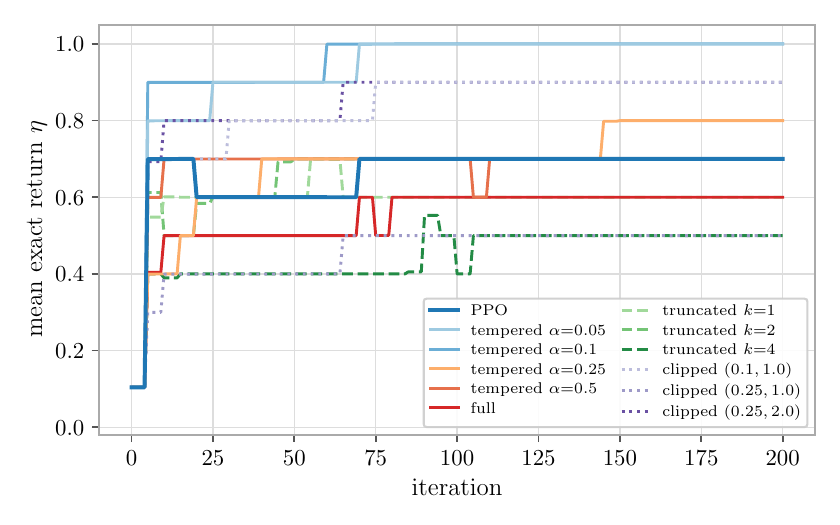}
\caption{Mean exact return over $10$ seeds versus iteration for the $793$k-parameter transformer under N-SGD ($\mathtt{lr}=1.0$, $K=4$). Outcomes are bimodal ($\eta\approx0$ or $1$), so the mean is the success-rate curve; the success rates are in Table~\ref{tab:results-summary}.}
\label{fig:h-main}
\end{figure}

Same reward and PPO scaffold with the $793$k-parameter transformer, sample-based $\eta$ on fresh $1024$-trajectory batches, N-SGD, $\mathtt{lr} \in \{0.3, 1.0, 3.0\}$, $10$ seeds. At $\mathtt{lr} = 0.3$ all methods succeed $10/10$; at $3.0$ everything is too aggressive (best: tempered $\alpha = 0.25$ at $2/10$).

The discriminating rate is $\mathtt{lr} = 1.0$ (Figure~\ref{fig:h-main}, Table~\ref{tab:h-detail}), where the interior tempered variants succeed on every seed and PPO on seven of ten. Losing seeds fail by self-reinforcing wrong-branch commitment --- $P(y_0 \in \{\mathtt{B}, \mathtt{D}\}) \to 1$ within iterations $5$--$10$, after which no batch contains a rewarded trajectory and the gradient estimate is zero.

\begin{table}[h]
\centering
\small
\begin{tabular}{lcc}
\toprule
Method & Success rate ($n = 10$) & Final $\eta$ (mean $\pm$ SEM) \\
\midrule
\textbf{Tempered $\alpha = 0.05$} & $\mathbf{10/10}$ & $\mathbf{1.000 \pm 0.000}$ \\
\textbf{Tempered $\alpha = 0.10$} & $\mathbf{10/10}$ & $\mathbf{1.000 \pm 0.000}$ \\
Clipped $(0.10,\,1.0)$           & $9/10$           & $0.900 \pm 0.100$ \\
Clipped $(0.25,\,2.0)$           & $8/9$            & $0.889 \pm 0.111$ \\
Tempered $\alpha = 0.25$         & $8/10$           & $0.800 \pm 0.133$ \\
PPO                              & $7/10$           & $0.700 \pm 0.153$ \\
Tempered $\alpha = 0.50$         & $7/10$           & $0.700 \pm 0.153$ \\
Truncated $k = 2$                & $7/10$           & $0.700 \pm 0.153$ \\
Full                             & $6/10$           & $0.600 \pm 0.163$ \\
Truncated $k = 1$                & $6/10$           & $0.600 \pm 0.163$ \\
Truncated $k = 4$                & $5/10$           & $0.500 \pm 0.167$ \\
Clipped $(0.25,\,1.0)$           & $5/10$           & $0.500 \pm 0.167$ \\
\bottomrule
\end{tabular}
\caption{Transformer under N-SGD at $\mathtt{lr} = 1.0$ ($K = 4$, $10$ seeds): success rate and final $\eta$ per variant.}
\label{tab:h-detail}
\end{table}

\subsection{Transformer under Adam}
\label{app:i}

Same setup with Adam, $\mathtt{lr} \in \{10^{-3}, 3 \times 10^{-3}, 10^{-2}\}$. Adam's per-parameter RMS normalization partially absorbs gradient-direction bias (Appendix~\ref{app:opt-ablation}).

\begin{figure}[h]
\centering
\includegraphics[width=0.7\linewidth]{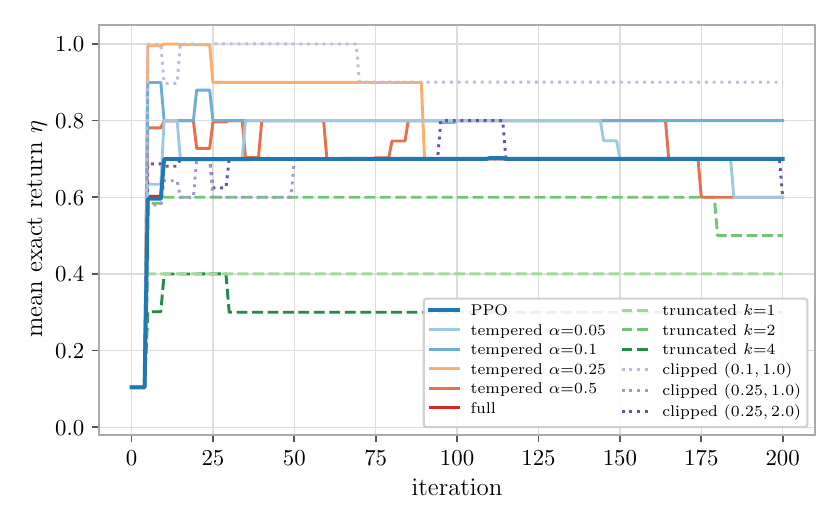}
\caption{Adam ablation of Figure~\ref{fig:h-main} ($\mathtt{lr} = 3 \times 10^{-3}$, $K = 4$, $10$ seeds). The success-rate advantage survives at reduced amplitude (Table~\ref{tab:i-detail}).}
\label{fig:i-detail}
\end{figure}

At $\mathtt{lr} = 10^{-3}$ all methods succeed $10/10$. At $3 \times 10^{-3}$ (Figure~\ref{fig:i-detail}, Table~\ref{tab:i-detail}) clipped $(0.1, 1.0)$ and tempered $\alpha = 0.1$ lead PPO by $19$~pp, roughly two thirds of the N-SGD gap. At the aggressive $10^{-2}$ (Table~\ref{tab:i-aggressive}) tempered $\alpha = 0.1$ keeps a $20$~pp lead over PPO while the full correction collapses to no successful seed, the worst method: Adam's per-parameter adaptivity applied to the accumulated ratio produces unbounded updates on outlier tokens, the failure mode CTPO's $\sqrt{t}$-scaled clip prevents.

\begin{table}[h]
\centering
\small
\begin{tabular}{lcc}
\toprule
Method & Success rate & Final $\eta$ (mean $\pm$ SEM) \\
\midrule
\textbf{Clipped $(0.10,\,1.0)$} & $\mathbf{8/9}$   & $\mathbf{0.889 \pm 0.111}$ \\
Tempered $\alpha = 0.10$        & $8/10$           & $0.800 \pm 0.133$ \\
Clipped $(0.25,\,1.0)$          & $7/9$            & $0.778 \pm 0.147$ \\
PPO                             & $7/10$           & $0.700 \pm 0.153$ \\
Tempered $\alpha = 0.25$        & $7/10$           & $0.700 \pm 0.153$ \\
Full                            & $7/10$           & $0.700 \pm 0.153$ \\
Tempered $\alpha = 0.05$        & $6/10$           & $0.600 \pm 0.163$ \\
Tempered $\alpha = 0.50$        & $6/10$           & $0.600 \pm 0.163$ \\
Truncated $k = 2$               & $5/9$            & $0.556 \pm 0.176$ \\
Clipped $(0.25,\,2.0)$          & $5/9$            & $0.556 \pm 0.176$ \\
Truncated $k = 1$               & $4/10$           & $0.400 \pm 0.163$ \\
Truncated $k = 4$               & $3/9$            & $0.333 \pm 0.167$ \\
\bottomrule
\end{tabular}
\caption{Transformer under Adam at $\mathtt{lr} = 3 \times 10^{-3}$ ($K = 4$): success rate and final $\eta$ per variant.}
\label{tab:i-detail}
\end{table}

\begin{table}[h]
\centering
\small
\begin{tabular}{lcc}
\toprule
Method & Success rate & Final $\eta$ (mean $\pm$ SEM) \\
\midrule
\textbf{Tempered $\alpha = 0.10$} & $\mathbf{5/10}$ & $\mathbf{0.500 \pm 0.167}$ \\
PPO                              & $3/10$          & $0.300 \pm 0.153$ \\
Tempered $\alpha = 0.05$         & $2/10$          & $0.200 \pm 0.133$ \\
Tempered $\alpha = 0.25$         & $2/10$          & $0.200 \pm 0.133$ \\
Clipped $(0.25,\,1.0)$           & $2/9$           & $0.222 \pm 0.147$ \\
Clipped $(0.25,\,2.0)$           & $2/9$           & $0.222 \pm 0.147$ \\
Tempered $\alpha = 0.50$         & $1/10$          & $0.100 \pm 0.100$ \\
Truncated $k = 1$                & $1/10$          & $0.100 \pm 0.100$ \\
Clipped $(0.10,\,1.0)$           & $0/9$           & $0.002 \pm 0.002$ \\
Truncated $k = 2$                & $0/9$           & $0.000$ \\
Truncated $k = 4$                & $0/9$           & $0.000$ \\
\textbf{Full ($\alpha = 1$)}     & $\mathbf{0/10}$ & $\mathbf{0.000}$ (worst) \\
\bottomrule
\end{tabular}
\caption{Transformer under Adam at $\mathtt{lr} = 10^{-2}$ ($K = 4$): success rate and final $\eta$ per variant.}
\label{tab:i-aggressive}
\end{table}

\subsection{Optimizer-absorption ablation on the tabular policy}
\label{app:opt-ablation}

Swapping N-SGD for Adam on the tabular two-branch setup shrinks the interior-tempered-over-PPO advantage at iteration $50$ from $\eta = 0.961$ vs.\ $0.341$ under N-SGD to a relative lead of about $20\%$ --- Adam's per-parameter RMS normalization rescales the biased direction back toward the true gradient, absorbing part but not all of the estimator-level bias. The transformer result ($30$~pp $\to$ $19$~pp, Appendix~\ref{app:i}) matches this tabular prediction, completing the optimizer $\times$ parameterization $2 \times 2$ and showing the mechanism decomposition is not sensitive to parameterization scale at the sizes tested.

\section{Pretrained-LM study: structured-record consistency at \texorpdfstring{$0.5$B}{0.5B}}
\label{app:pretrained}

We test whether the gradient-and-training account of Section~\ref{sec:mechanism} predicts the LM-scale effect on Qwen2.5-0.5B (base). The reward is a rule-verifiable string check on structured JSON: the model fills a $6$-slot template (\texttt{status/service/region/retries/load/code}), each slot a masked $4$-way choice with all candidates single-token in context (tokenizer gate $24/24$), and the reward reads two slots. The complete-sequence space is $4^6 = 4096$, so $\eta(\pi)$, $S_1$ marginals, and $P(S_6 \mid S_1)$ are exactly enumerable at any training point.

\paragraph{Protocol.} Custom masked-slot generation and PPO update --- no TRL/HF-Trainer scaffold --- so all six variants share one code path differing only in the $g(L^s_t)$ switch (asserted by a bit-identical-loss test at $\alpha = 0$). Frozen config: $256$ rollouts/iteration, minibatch $8$, $K = 4$, $\epsilon = 0.2$, Adam ($\beta_1 = 0.9$, $\beta_2 = 0.95$), grad clip $1.0$, entropy $= 0$, ref-KL $= 0$ (no external restoring force), $\mathtt{lr} = 3 \times 10^{-6}$ frozen by pilot; advantage $=$ trajectory return minus batch-mean baseline. $10$ seeds per (cell, method), extended to $16$ paired seeds for the pre-registered headline pair (PPO vs.\ tempered $\alpha = 0.10$) in Cells 1, 3, 4. bf16 with gradient checkpointing on one L4.

\paragraph{Cell design.} All cells place reward at the terminal slot $S_6$. The reward and an $N = 8$ few-shot priming prefix vary whether terminal value depends on the earlier slot $S_1$ and whether the sampling loop is susceptible to a low-reward trap:

\begin{center}
\footnotesize
\resizebox{0.95\textwidth}{!}{%
\begin{tabular}{llcccl}
\toprule
Cell & Priming ($S_1$ in $N=8$) & Reward & $\eta_0$ & $\mathbb{E}[\text{rew}/256]$ & Role \\
\midrule
1 & NEUTRAL (uniform)                & TWO\_BR (matched)    & $0.386$ & $99$ & Prefix-dependent value; dense reward prevents a trap \\
3 & FATAL (unknown, soften $0.25$)   & TWO\_BR              & $0.033$ & $8.5$ & Prefix-dependent value; designed low-reward trap \\
4 & NEUTRAL (uniform)                & LATE\_ONLY           & $0.398$ & $102$ & Terminal reward independent of $S_1$ --- null control \\
\bottomrule
\end{tabular}}
\end{center}

Pre-run sign-off measurements (per-condition $S_1$ marginals, full $P(S_6 \mid S_1)$, $\eta_0$ per reward, off-policiness dry-run, $\alpha^\star = 1/(T\sigma^2)$ heuristic in $[0.02, 0.3]$ for every cell) are recorded in \texttt{FREEZE\_TABLE.md}; all $12$ pre-flight gates pass.

\paragraph{Pilot and endpoint revision.} A Cell-1 pilot ($2$ methods $\times$ $3$ lrs $\times$ $3$ seeds, $50$ iterations) froze $\mathtt{lr} = 3 \times 10^{-6}$ and showed a $+9$~pp early lead for tempered $\alpha = 0.10$ at iteration $3$, collapsing by iteration $10$ as both arms hit ceiling ($18/18$ pilot runs succeeded). The pilot revised the primary endpoint from iters-to-$\eta_{0.9}$ (which the $\mathtt{eval\_every} = 10$ discretisation cannot resolve for sub-$10$-iteration effects) to \emph{mean MC $\eta$ over iterations $2$--$8$}; Cell 3 kept the brief's original success-rate endpoint, since a sampling trap would turn mean-$\eta$ into a mixture of successful and failed runs. The pilot table is in \texttt{outputs/pilot/}.

\paragraph{Cell 1 --- rate advantage transfers.} At $\eta_0 = 0.386$ ($\approx 99$ rewarded trajectories per batch), a sampling trap is implausible and the analysis predicts a \emph{rate} advantage, which is what we observe (Table~\ref{tab:v2-cell1-rate}): tempered $\alpha = 0.10$ climbs faster over iterations $2$--$5$, collapsing to noise by iteration $10$ at ceiling.

\begin{table}[h]
\centering
\small
\caption{Cell 1 (NEUTRAL $N=8$, TWO\_BRANCH) per-iteration MC $\eta$, $16$ paired seeds. $16/16$ seeds positive at iteration $3$, $15/16$ at $5$; the effect reaches the noise floor by iteration $7$ as both arms hit ceiling.}
\label{tab:v2-cell1-rate}
\begin{tabular}{lcccccc}
\toprule
Method & iter 2 & iter 3 & iter 4 & iter 5 & iter 10 & iter 20 \\
\midrule
PPO           & $0.534$ & $0.756$ & $0.884$ & $0.957$ & $0.998$ & $0.999$ \\
Tempered $\alpha = 0.10$ & $\mathbf{0.706}$ & $\mathbf{0.910}$ & $\mathbf{0.978}$ & $\mathbf{0.994}$ & $0.999$ & $0.999$ \\
\midrule
$\Delta$ (T $-$ P) pp & $+17.1$ & $+15.4$ & $+9.4$ & $+3.7$ & $+0.1$ & $+0.05$ \\
\bottomrule
\end{tabular}
\end{table}

The full method grid (Table~\ref{tab:v2-cell1-grid}; all $72$ Cell-1 runs succeed, so the cell is purely rate-mode) reproduces the toy interior-$\alpha$ profile, peaking at $\alpha = 0.10$ and collapsing at $\alpha = 1$ where the variance penalty cancels the bias correction. The peak brackets the pre-registered $\alpha^\star \approx 0.08$ heuristic; clipped $(0.10, 1.0)$ is indistinguishable from tempered $0.10$, as expected when the log-space clip rarely binds at this KL scale.

\begin{table}[h]
\centering
\small
\caption{Cell 1 full grid: paired difference vs.\ PPO on mean MC $\eta$ over iterations $2$--$8$ (the pre-registered rate endpoint); normal-approximation $95\%$ CI on the paired mean.}
\label{tab:v2-cell1-grid}
\begin{tabular}{lcccc}
\toprule
Method & $n$ (paired seeds) & $\Delta$ vs.\ PPO (pp) & seeds positive & $95\%$ CI (pp) \\
\midrule
Tempered $\alpha = 0.05$ & $10$ & $+4.5$ & $9/10$  & $\pm 2.6$ \\
Tempered $\alpha = 0.10$ & $16$ & $\mathbf{+6.7}$ & $\mathbf{16/16}$ & $\pm 2.5$ \\
Tempered $\alpha = 0.25$ & $10$ & $+5.8$ & $10/10$ & $\pm 3.4$ \\
Clipped $(0.10,\,1.0)$   & $10$ & $+6.5$ & $10/10$ & $\pm 3.4$ \\
Full ($\alpha = 1$)      & $10$ & $+0.1$ & $6/10$  & $\pm 1.9$ \\
\bottomrule
\end{tabular}
\end{table}

\paragraph{Cell 3 --- the success-rate mechanism does not fire.} At $\eta_0 = 0.033$ ($\mathbb{E}[\text{rew}/256] = 8.5$, just above the absorbing floor), a sampling trap would require the biased-gradient direction to drive the policy toward the fatal-primed marginal in iterations $1$--$5$; if it wins, $\eta$ falls below $0.020$, zero-reward batches become likely ($\ge 5.7 \times 10^{-3}$ per iteration, cumulatively $\gtrsim 10\%$ over $20$), and the run pins.

The measured per-iteration MC fatal marginal \emph{does not rise on any of the $16$ PPO seeds}: it starts in $[0.31, 1.00]$ (Poisson-noised around the exact $0.742$ under FATAL priming) and declines monotonically, reaching mean $\le 0.02$ by iteration $10$ and $\le 10^{-3}$ by $15$ (Table~\ref{tab:v2-cell3-fatal-drift}). One seed (\texttt{PPO\_s2}) plateaus near $0.98$ for iterations $1$--$5$ before collapsing --- the expected signature at a high-initial-fatal seed where rewarded rollouts number $1$--$2$ per batch --- but the biased direction never dominates. All $16$ PPO seeds succeed; tempered $\alpha = 0.10$ shows the same pattern ($16/16$; highest initial fatal marginal $0.996$, collapsing to $0.000$ by iteration $5$).

\begin{table}[h]
\centering
\small
\caption{Cell 3 per-iteration fatal-marginal decline (mean/max/min over $16$ PPO seeds). No seed rises; all reach $\le 10^{-2}$ by iteration $10$.}
\label{tab:v2-cell3-fatal-drift}
\begin{tabular}{lccccccccc}
\toprule
iter & 1 & 2 & 3 & 4 & 5 & 7 & 10 & 15 & 20 \\
\midrule
mean & 0.778 & 0.649 & 0.454 & 0.333 & 0.229 & 0.096 & 0.014 & 0.001 & 0.000 \\
max & 0.996 & 0.996 & 0.988 & 0.957 & 0.973 & 0.668 & 0.180 & 0.008 & 0.004 \\
min & 0.312 & 0.141 & 0.035 & 0.012 & 0.000 & 0.000 & 0.000 & 0.000 & 0.000 \\
\bottomrule
\end{tabular}
\end{table}

The reason is structural: in the toy MDP a wrong first-token commitment yields \emph{deterministically} zero reward under any continuation, whereas at LM scale $\approx 8.5$ rewarded trajectories per batch arrive in expectation even under FATAL priming, and their advantage signal dominates the biased-gradient drift from iteration $1$.

\emph{Truncation disclosure.} Cell 3's non-headline arms (tempered $\alpha \in \{0.05, 0.25\}$, clipped, full) were cancelled after the PPO tier completed $16/16$ at ceiling with no fatal-marginal drift --- no success-rate comparison was possible at ceiling. Only the pre-registered headline pair ran to $16$ seeds; the other methods are retained in Cell 1's grid, where the rate endpoint discriminates.

\paragraph{Cell 4 --- the pre-registered falsifier passes.} Cell 4 keeps Cell 1's priming (same NEUTRAL $N = 8$ lines, same initial $S_1$ marginal) and swaps only the reward: LATE\_ONLY $= \mathbf{1}[S_6 = \text{target}]$ pays regardless of $S_1$, removing reward dependence on $S_1$ while holding terminal reward placement and the off-policiness scale ($\mathrm{Var}(\log R^s_5) = 10.4$, identical to Cell 3's gate value) fixed. The pre-registered prediction is $\Delta \eta \approx 0$; a Cell-1-sized effect here would contradict the Cell 1 attribution.

\begin{table}[h]
\centering
\small
\caption{Cell 4 (NEUTRAL $N=8$, LATE\_ONLY) per-iteration MC $\eta$, same protocol and $16$ paired seeds as Table~\ref{tab:v2-cell1-rate}. Per-seed signs at iterations $2$--$5$: $8/16$, $9/16$, $8/16$, $5/16$ positive --- at or below chance, against Cell 1's $16/16$ at iterations $2$--$3$.}
\label{tab:v2-cell4-null}
\begin{tabular}{lcccccc}
\toprule
Method & iter 2 & iter 3 & iter 4 & iter 5 & iter 10 & iter 20 \\
\midrule
PPO           & $0.663$ & $0.876$ & $0.955$ & $0.985$ & $0.999$ & $0.999$ \\
Tempered $\alpha = 0.10$ & $0.715$ & $0.921$ & $0.984$ & $0.994$ & $0.998$ & $0.999$ \\
\midrule
$\Delta$ (T $-$ P) pp & $+5.1$ & $+4.6$ & $+2.9$ & $+0.9$ & $-0.05$ & $-0.02$ \\
\bottomrule
\end{tabular}
\end{table}

On the primary endpoint the paired difference is $+1.9$~pp ($95\%$ CI $\pm 1.8$~pp) with $8/16$ seeds positive --- a two-sided sign test at exactly chance ($p = 1.0$), against Cell 1's $+6.7$~pp at $16/16$ ($p \approx 3 \times 10^{-5}$): a $3.5\times$ attenuation in mean and a complete collapse in per-seed consistency.

The paired-mean CI marginally excludes zero, and the residual has a mechanistic home --- unlike the tabular late-only null, where the optimum provably leaves early tokens uniform, the pretrained conditional $P(S_6 \mid S_1)$ is not exactly $S_1$-independent, so the reward-relevant prefix dependence is attenuated rather than eliminated exactly at LM scale. The per-seed $S_1$ endpoints corroborate this: Cell 1's PPO seeds all commit $S_1$ to a reward-matched branch by iteration $20$, while Cell 4's scatter across seed-dependent directions (two seeds drift onto \texttt{unknown}, a value TWO\_BRANCH would kill) --- free drift under no restoring force, not a reward-required commitment.

The full Cell 4 grid sharpens the null into a positive prediction: with no bias to correct, the correction's \emph{cost} should remain. The full correction is strictly worse than PPO --- $-4.4$~pp, $0/10$ seeds positive, $95\%$ CI $\pm 2.5$~pp --- the variance penalty with no compensating bias reduction, mirroring the late-only control in the reward ladder of Section~\ref{sec:setup} and V-trace's motivation for refusing the accumulated product. Interior variants sit at small, sign-inconsistent residuals (tempered $\alpha = 0.05$: $+1.2$~pp at $5/10$; $\alpha = 0.25$: $+1.1$~pp at $4/10$; clipped: $+2.6$~pp at $6/10$). All $72$ Cell-4 runs succeed, as expected at $\mathbb{E}[\text{rew}/256] = 102$.

\section{Scheduling: nulls, controls, and boundary}
\label{app:scheduling}

\subsection{With no measured drift, the correction only costs variance}
\label{app:sched-driftdial}

At realistic operating points a held correction is null to harmful, for two measurable reasons (Table~\ref{tab:sched-null}). The first is that realistic training accumulates no staleness to repair. At published FJSP hyperparameters the drift after a full $K = 3$ update is $\sigma^2 T \approx 3 \times 10^{-5}$, and the correction loses to PPO in every cell, winning $4$ of $60$ paired comparisons (Appendix~\ref{app:sched-driftdial}). With no stale state distribution to repair, the credit-routing channel pays its horizon-dependent variance cost and buys nothing back.

The second reason is why raising drift alone does not help. Dialing the learning rate, clip, and $K$ until the drift rises by five orders of magnitude makes the paired final on-time difference cross zero, with every paired seed won at the top, but only at a deadline where training sits at the reward floor; at a fair deadline the same dial yields a null, because with plentiful signal PPO relearns from fresh batches on its own. Neither a richer reward nor a drifting learner rescues the held correction: a rich-reward benchmark built to give the correction its best case is null at the largest drift measured in this paper (Appendix~\ref{app:sched-driftdial}), and a drifting transformer policy under a scarce reward loses by a small margin at every deadline, turning positive in sign only when the update is so aggressive that PPO itself diverges (Appendix~\ref{app:sched-transformer}). The two failures are independent: where drift is realistic there is nothing to repair, and where the reward retains signal, PPO absorbs the staleness on its own, whatever the drift.

\begin{table}[t]
\centering
\footnotesize
\begin{tabular}{lcccc}
\toprule
Setting & $\sigma^2 T$ & Endpoint & $\Delta$ (corr.\ $-$ PPO) & Seed wins \\
\midrule
\multicolumn{5}{l}{\emph{HGNN dispatcher, FJSP $10 \times 5$, $T = 50$; correction held for the whole run}} \\
Published regime ($K = 3$), two deadlines & $3 \times 10^{-5}$ & final on-time & $-6.0$, $-8.6$~pp & $4/60$ \\
Drift dial, hard deadline (reward floor) & $1.3$  & final on-time & $-5.2$~pp & $1/5$ \\
                                                & $15$   &               & $+4.4$~pp & $3/5$ \\
                                                & $34$   &               & $+9.0$~pp & $5/5$ \\
Drift dial, fair deadline (saturated) & $1.7$  & final on-time & $+1.8$~pp & $3/5$ \\
                                                & $7.2$  &               & $-3.8$~pp & $3/5$ \\
                                                & $22.5$ &               & $-3.2$~pp & $2/5$ \\
Rich graded reward (resting windows) & $0.7$  & AUC & $-0.080$ & $1/5$ \\
                                                & $38.8$ &     & $-0.039$ & $2/5$ \\
                                                & $60.7$ &     & $-0.027$ & $3/5$ \\
Acquisition, hard deadline, fresh seeds & $34$   & early AUC & $-0.019$ & $8/16$ \\
\midrule
\multicolumn{5}{l}{\emph{Transformer pointer policy, three-machine bottleneck flow shop, $J = 12$; correction held}} \\
Dense makespan reward                           & $1.38$ & AUC & n.s.     & $9/16$ \\
Tight deadline, floor $0.14$ (scarce) & $1.38$ & AUC & $-0.062$ & $5/16$ \\
\quad floor $0.35$                              &        &     & $-0.034$ & $6/16$ \\
\quad floor $0.56$                              &        &     & $-0.012$ & $5/15$ \\
Update dial D1 (PPO final $0.95$) & ---    & AUC & $-0.067$ & $7/16$ \\
\quad D2 (PPO final $0.47$)                     & ---    &     & $-0.045$ & $8/16$ \\
\quad D3 (PPO final $0.30$)                     & ---    &     & $+0.122$ & $9/16$ \\
\bottomrule
\end{tabular}
\caption{Held-correction results at and beyond realistic operating points. $\Delta$ is the paired, per-seed difference (corrected minus PPO) on the stated endpoint; seed wins count paired seeds favoring the correction. Drift $\sigma^2 T$ is measured on the rollout policy after one full $K$-epoch update, before training, and was not recorded for the transformer update-strength dial. Details, the sizing rule for $\alpha$, and the registered endpoints are in Appendices~\ref{app:sched-driftdial} and~\ref{app:sched-transformer}.}
\label{tab:sched-null}
\end{table}

\paragraph{Empirical test: the drift dial.} We tested the correction on flexible job-shop scheduling (FJSP, $10 \times 5$, binary on-time terminal reward $\mathbf{1}[C_{\max} \le c \cdot C^*]$, HGNN dispatching policy of \citet{song2022flexible}, horizon $T = 50$), in both the free-order (DAG) protocol and the canonical-order (tree) protocol of Appendix~\ref{app:derivation}. At the published training regime (lr $2 \times 10^{-4}$, clip $0.2$, $K = 3$ reuse epochs) the correction \emph{loses} to PPO in every cell of both protocols; Table~\ref{tab:sched-null} lists the canonical-order cell at $\alpha = 0.03$ and the count of paired comparisons won. A direct measurement says why: the per-step log-ratio variance between the data-collecting and updated policy, measured \emph{after a full} $K$-epoch update, is $\sigma^2 \approx 5 \times 10^{-7}$, i.e.\ a prefix log-ratio of magnitude $\approx 0.005$ over the whole episode. The staleness that the correction reweights for is absent --- the omitted term is measured to be $\approx 0$ --- while the credit-routing channel still pays its $\alpha \cdot T$ variance cost on a sparse binary signal.

\begin{figure}[t]
\centering
\includegraphics[width=0.62\linewidth]{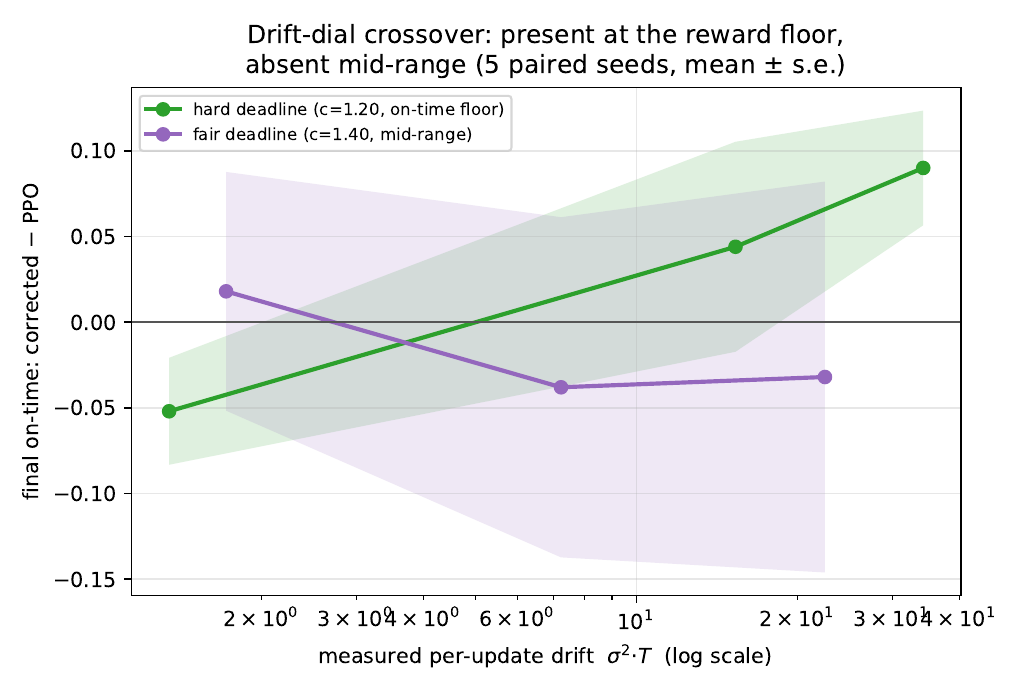}
\caption{\textbf{The drift-dial on canonical-order FJSP, in two reward regimes.} Paired final on-time difference (corrected minus PPO; $5$ paired seeds, mean $\pm$ s.e.) against the drift measured at each training regime before launching, $\sigma^2 T$ (log scale). At the hard deadline (green), where training operates at the on-time floor, the gap is monotone in measured drift and crosses zero, with every paired seed won at the top on both registered endpoints. At the fair deadline (purple), where all runs train mid-range, the same dial yields a null: with plentiful signal PPO absorbs the staleness on its own. The published training regime lies off the left edge, where the correction loses everywhere: with nothing stale to correct, only its variance cost remains. Values in Table~\ref{tab:sched-null}.}
\label{fig:drift-crossover}
\end{figure}

We therefore turned the drift dial --- raising the learning rate $10$--$30\times$, widening or removing the clip, and setting $K = 12$ --- producing three settings whose measured drifts (Table~\ref{tab:sched-null}) sit five to six orders of magnitude above the published regime; PPO's own performance falls from $0.96$ to $0.05$--$0.11$, confirming that the staleness bias is real. At each setting the correction strength was frozen \emph{before} training by the rule $\alpha = \min(1,\, 1/(T\sigma^2))$ from the measured drift, and both arms share all hyperparameters and paired seeds. The registered readout, the paired final on-time difference, is monotone in measured drift and crosses zero, with every paired seed won at the highest drift (Table~\ref{tab:sched-null}; Figure~\ref{fig:drift-crossover}); the pre-registered continuous secondary endpoint, mean makespan gap to $C^*$, agrees there on every seed ($0.342$ vs.\ $0.395$). The absolute on-time levels at these settings sit near the floor ($0.05$--$0.15$), so we read the result as a sign-and-ordering finding rather than an effect-size estimate. At the lowest drift the sizing rule licenses a near-full correction ($\alpha = 0.74$), which loses.

A registered follow-up retested the same three training regimes away from the reward floor, at a deadline recalibrated (by a PPO-only pilot, blind to the corrected method) so that all settings train into the $0.55$--$0.95$ on-time range; the drift re-measured at that deadline is lower (Table~\ref{tab:sched-null}). There the result is a \emph{null}: all three paired differences are within seed noise, the secondary endpoint agrees, and the strong-correction arm no longer loses either ($\alpha = 0.58$ at the lowest drift). Figure~\ref{fig:drift-crossover} overlays both regimes: with plentiful reward signal, PPO recovers from stale data within a few updates on its own.

\begin{figure}[t]
\centering
\includegraphics[width=0.95\linewidth]{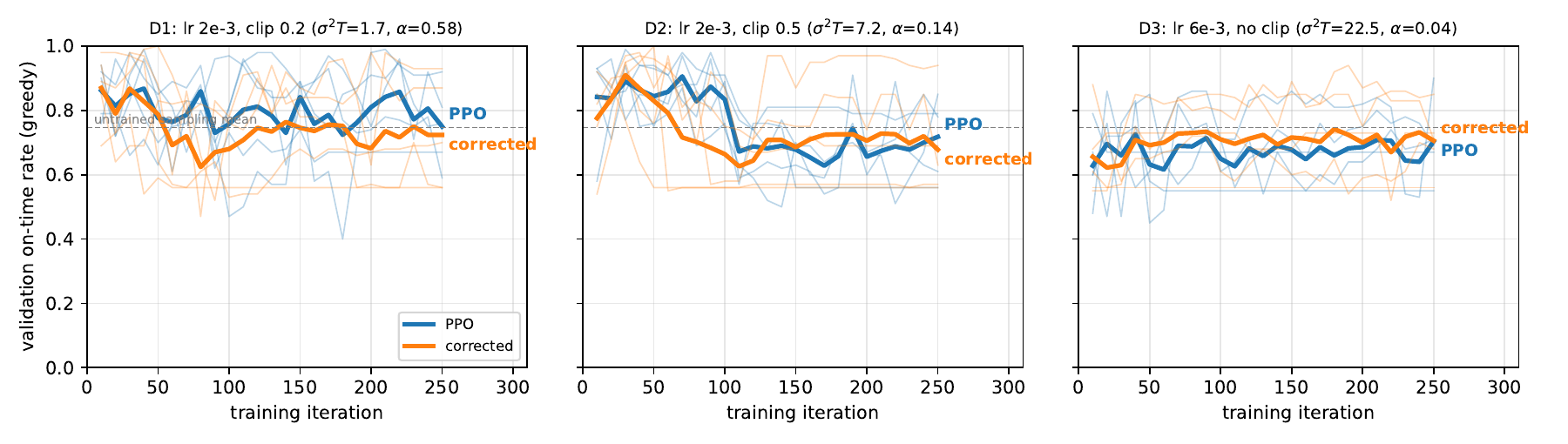}
\caption{\textbf{Reward saturation at the fair deadline (early training, iterations $0$--$250$).} Validation on-time rate at the recalibrated deadline for the three drift settings ($5$ paired seeds thin, mean bold; dashed line: the untrained sampling policy's mean success rate, $0.75$). There is no extended learning phase: both methods reach their operating level within tens of iterations --- near or below the untrained baseline --- and under the drift settings decay from an early peak at matched rates. The binary indicator is blind to the remaining, still-large gap to optimal makespans ($\approx 0.35$ mean), so the fair-deadline comparison resolves seed noise around a saturated reward rather than a difference in learning.}
\label{fig:easy-early}
\end{figure}

The fair-deadline null itself carries a caveat that bounds how much either method could have shown there: the binary reward \emph{saturates at initialization}. At the recalibrated deadline the untrained sampling policy already succeeds on $75\%$ of instance--samples (per-instance range $0.27$--$0.97$), and the training curves (Figure~\ref{fig:easy-early}) show no extended learning phase --- both methods reach their operating level within the first $\sim\!10$--$30$ iterations, peak early, and under the drift settings \emph{erode} from that peak at matched rates rather than climb. Meanwhile the continuous quality metric shows both plateauing at a mean makespan gap of $\approx 0.35$ above optimal, versus $\approx 0.13$ reachable at published hyperparameters: the indicator reward stops registering schedule quality long before quality is good, so past the first few updates it provides no gradient toward better schedules for either method to exploit. The binary on-time reward in this problem class thus has a narrow informative band --- starved at the hard deadline under drift, saturated at the fair one --- and endpoint differences at the fair deadline are dominated by seed noise around an early-attained plateau.

\paragraph{A rich-reward benchmark built for the correction --- and a null.}
A third experiment gives the correction its best case with an \emph{informative} signal, to test whether the fair-deadline null was an artifact of the indicator reward:
FJSP with resting-time windows (a hard minimum lag between consecutive
operations of a job, enforced by the environment; a soft maximum lag whose
violation is measured), and a graded terminal reward
$R = \mathbf{1}[V{=}0] - (C_{\max}-C^*)/C^* - V/C^*$ combining a
feasibility bonus, the normalized makespan gap, and the normalized
violation sum $V$. Missing a window is a \emph{delayed} consequence of
early machine-allocation decisions --- the credit-routing structure the correction targets. The benchmark was difficulty-calibrated
by a registered two-stage pilot (untrained feasibility in $[0.05,0.40]$;
then PPO-only learnability and headroom at 200 iterations), and the
confirmatory run registered sample efficiency (paired per-seed AUC of
greedy validation reward) as the primary endpoint, with $\alpha$ frozen
from probes measuring the two largest drifts of the project, plus a third
setting pre-declared as a no-drift control (Table~\ref{tab:sched-null}).
The result is a null across the board: the paired AUC difference is negative
at all three settings, nowhere near the registered criteria, with final-level
secondaries agreeing.

\paragraph{No acquisition-phase advantage even under scarce signal.}
An exploratory cut of the hard-deadline data suggested that the held correction also learns faster in the starved regime (corrected
arms at ${\sim}2\times$ PPO's on-time rate within the first $50$
iterations at the highest-drift setting; $4/5$ seeds, $+0.18$ early-phase
AUC, short of significance). A registered confirmation on $16$ fresh
paired seeds --- everything else frozen verbatim: pool, deadline,
hyperparameters, $\alpha = 0.03$ --- returned seed wins at chance and a mean
slightly below zero (Table~\ref{tab:sched-null}); the fresh seeds' PPO-only early level itself differed from the original seeds' by a factor of two. The held correction has no confirmed acquisition-phase advantage.

Among the held-correction experiments of this subsection, the correction's demonstrated value is survival under scarce signal with stale data: holding an already-learned policy against drift erosion; the warmup schedule of Appendix~\ref{app:sched-win} is a different device. Under the operating-point hypothesis of Section~\ref{sec:disc}, these nulls sit three orders of magnitude below the testbed in $N/T^2$ (Eq.~\ref{eq:peak-advantage}).

\subsection{Positive controls: the mechanism and its implementation are sound}
\label{app:sched-controls}

\begin{figure}[t]
\centering
\includegraphics[width=0.72\linewidth]{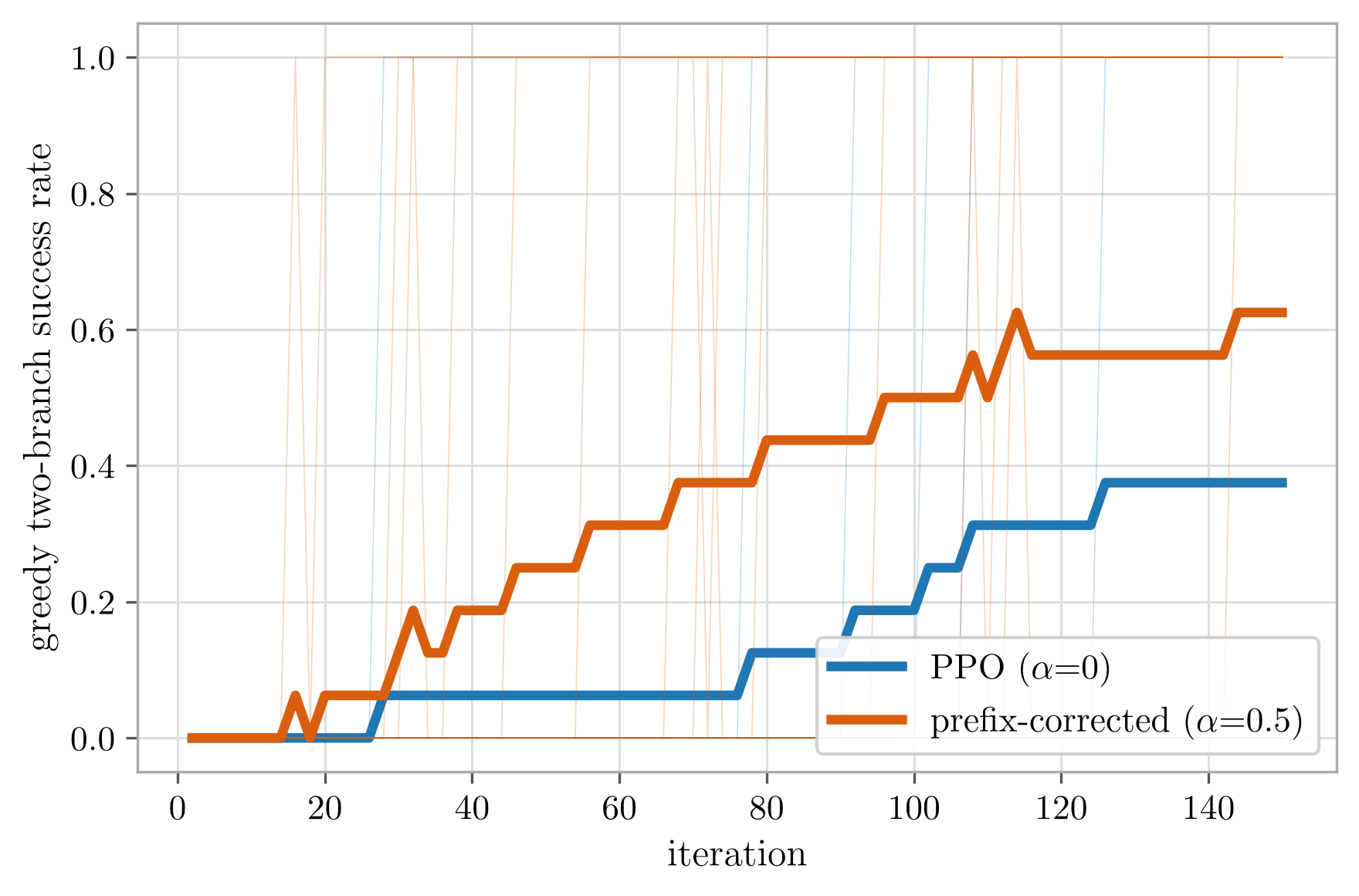}
\caption{\textbf{Free-choice positive control.} Greedy two-branch success vs.\ iteration under the LM-toy structure, the real correction code, N-SGD, and the FATAL dead-branch regime; $16$ seeds thin, mean bold; plain PPO ($\alpha=0$, blue) and the prefix correction ($\alpha=0.5$, orange). Primed toward a dead first branch, PPO frequently stays trapped at the reward floor while the correction routes credit back to the first action and escapes. Statistics in Table~\ref{tab:envladder}, first row.}
\label{fig:freecheck}
\end{figure}

\paragraph{Free-choice positive control.}
A control rules out a bug in the correction implementation as the cause of the scheduling nulls. It runs the \emph{same} correction code (\texttt{correction\_factor}, $\alpha=0$ recovering PPO bit-identically) on an environment stripped to the LM/toy structure: $T$ free $K$-way choices with the two-branch reward read from the first and last \emph{chosen} actions, a small policy that must \emph{learn} the prefix dependence (bag-of-prefix embeddings, no direct index on the branch symbol), and a GRPO group-relative baseline. Two ingredients decide whether the correction's benefit appears, both established elsewhere in this paper: the optimizer must be N-SGD (Adam absorbs part of the effect; Appendix~\ref{app:training}), and the run must sit in the dead-branch escape regime, primed toward a dead branch it must escape. In that regime the correction \emph{rescues runs from the absorbing dead-branch trap}, with several full rescues and no seed harmed (Table~\ref{tab:envladder}, first row; Figure~\ref{fig:freecheck}).

The control isolates the boundary. The same code reproduces the win the moment the environment and learner match the toy, so the scheduling non-reproduction lies in the environment (a coupled, constraint-driven dispatch process whose terminal outcome, the makespan-determining job, is emergent rather than chosen) or in the learner (a graph network with a value critic under Adam, rather than a sequence model under N-SGD with a group baseline), not in the mechanism or its implementation.

\paragraph{Env-only ablation.}
We vary \emph{only} the environment toward scheduling while freezing the learner,
optimizer, group baseline, FATAL dead-branch regime, and every hyperparameter, to
ask which scheduling property (if any) removes the win. Two env
modifications move the free-choice task toward scheduling: an
\emph{emergent} terminal, where the reward reads the modal symbol over
the whole trajectory (the analogue of the makespan-determining job)
rather than a directly chosen last token; and \emph{coupling}, where a
chosen symbol is masked as ``busy'' for a fixed number of steps so the
action set shrinks and changes, the constraint that prevents freely
piling work onto the target. Table~\ref{tab:envladder} reports the
paired advantage (sign test on non-tied seeds) at each rung.

\begin{table}[h]
\centering
\small
\begin{tabular}{llccc}
\toprule
env (learner/optimizer fixed) & size & PPO AUC & corr.\ $\Delta$AUC & sign test \\
\midrule
free (chosen terminal) & $T{=}8$ & $0.16$ & $+0.19$ & $10/0$, $p{=}0.001$ \\
emergent (whole-traj.\ terminal) & $T{=}8$ & $0.13$ & $+0.11$ & $8/1$, $p{=}0.02$ \\
emergent, longer horizon & $T{=}12$ & $0.07$ & $\mathbf{+0.21}$ & $15/0$, $p{<}10^{-4}$ \\
coupled (busy $=2$) & $T{=}8$ & $0.28$ & $+0.10$ & $10/3$, $p{=}0.05$ \\
coupled, strong (busy $=3$) & $T{=}8$ & $0.46$ & $-0.03$ & $8/11$, $p{=}0.82$ \\
\bottomrule
\end{tabular}
\caption{Env-only ablation, all else fixed. The correction's win survives
an emergent terminal and coupling, and \emph{grows} with a longer horizon
(deeper credit-routing chain). It vanishes only under strong coupling ---
where PPO's own score jumps to $0.46$ because forced exploration dissolves
the dead-branch trap: the task became easy, not the correction broken. The
advantage tracks the difficulty of the escape/credit-routing problem, not
the environment's ``scheduling-ness''.}
\label{tab:envladder}
\end{table}

\begin{figure}[t]
\centering
\includegraphics[width=\linewidth]{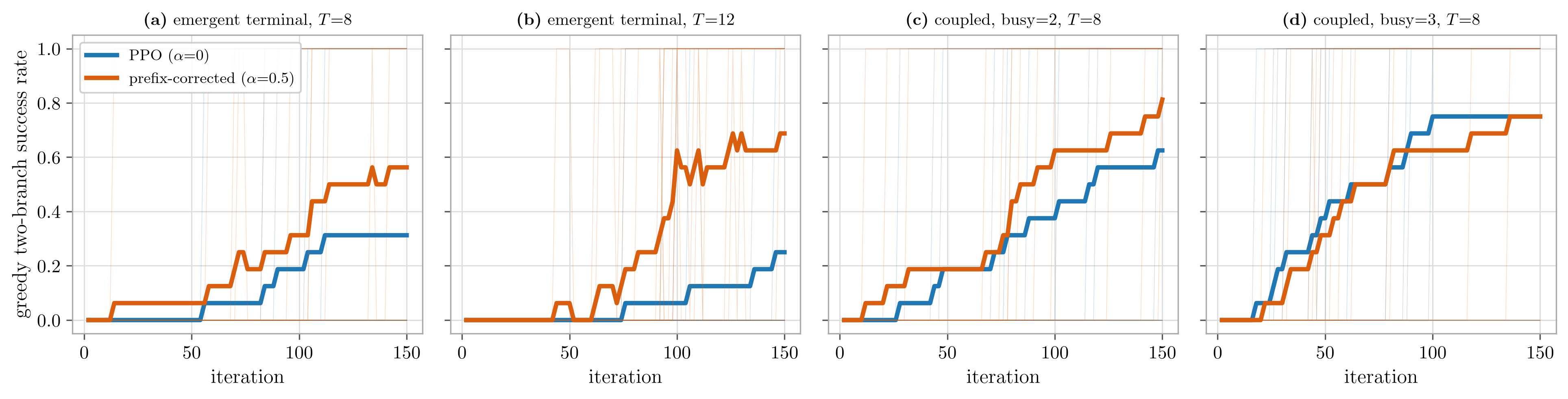}
\caption{\textbf{Env-only ablation learning curves} (learner, optimizer, regime, and hyperparameters all fixed; $16$ seeds thin, mean bold; greedy two-branch success). The prefix correction (orange) stays above PPO (blue) under an emergent terminal (panel~1), an emergent terminal over a longer horizon (panel~2), and coupling (panel~3); only strong coupling (panel~4) erases the gap, where both methods rise together. Statistics in Table~\ref{tab:envladder}.}
\label{fig:envladder}
\end{figure}

The advantage tracks whether a hard credit-routing problem is preserved, not emergent outcomes or coupling per se. With the correction code and the environment structure both excluded, the learner ($T = 50$ graph network) and optimizer (Adam with a value critic, low measured drift) remain as the operative cause, consistent with the $N/T^2$ prefactor and the optimizer ablations.

\subsection{The reward boundary: why the win does not transfer to realistic objectives}
\label{app:sched-boundary}

\paragraph{Mapping the boundary.}
Holding the learner and optimizer at the winning configuration
(N-SGD, group baseline), we asked directly what \emph{reward}
recovers the win on a scheduling task, and mapped the boundary by
construction. A single-machine on-time objective is greedily solvable
(Jackson's rule) and yields nothing; a two-machine flow-shop with
time-lags and varying, \emph{observed} instances is null because an observing policy reads off which jobs are critical and applies a
priority rule. Hiding the criticality does not recover it either --- and an all-jobs-on-time reward makes the feasible set
exponentially small, so both methods starve equally (the correction needs
\emph{some} rewarded trajectories to route credit from). Decomposing the
winning two-branch reward pinpoints why: it is sharp on \emph{one}
directly chosen early action whose ``liveness'' the policy must
\emph{discover}, and dense given that action --- a hard cliff followed by
a broad rewarded basin, with untrained success $\approx 0.5 \times 0.25 =
0.125$.

\begin{figure}[t]
\centering
\includegraphics[width=\linewidth]{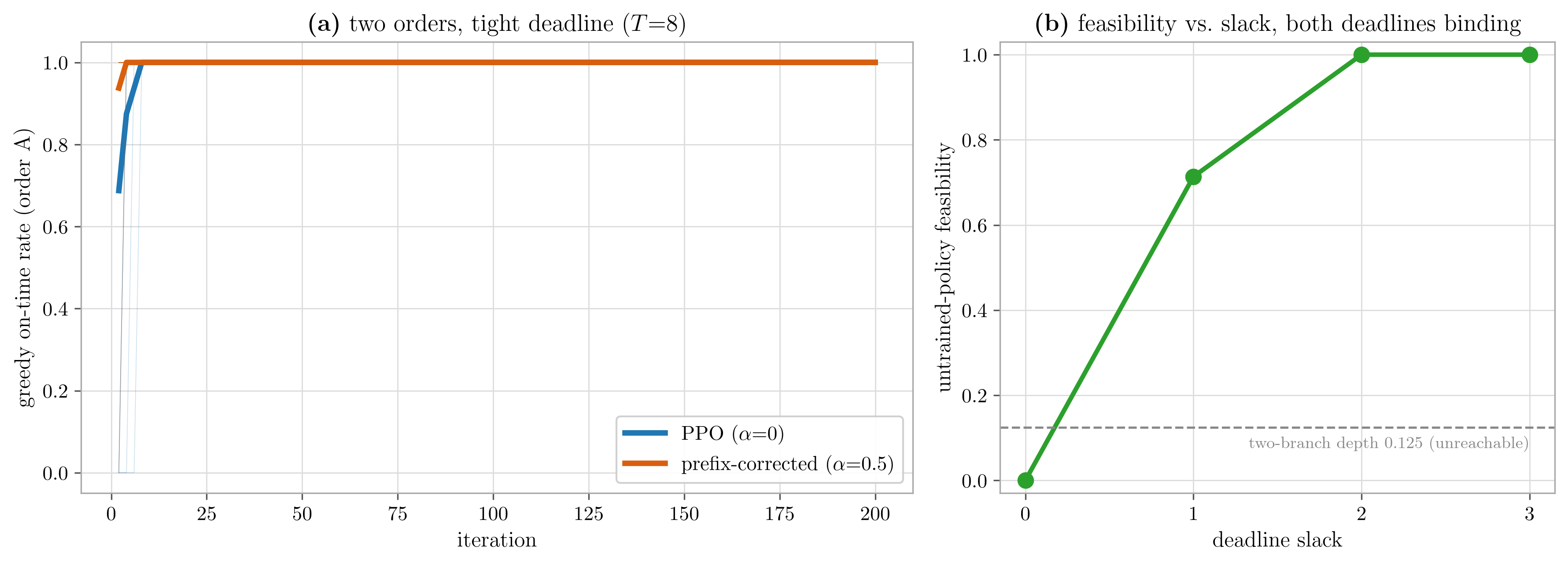}
\caption{\textbf{The reward boundary (learner/optimizer fixed at the
winning configuration).} \textbf{(A)} Two orders with a tight deadline ---
the closest scheduling analogue of the two-branch reward. The correction
(orange) beats PPO (blue) directionally and consistently (paired
$\Delta$AUC $+0.004$, $5/0$ non-tied seeds, $p=0.03$, never harmful): the
\emph{structure and sign} of the win transfer. \textbf{(B)} Adding a
second binding condition (both orders on tight deadlines) to deepen the
trap: the feasible fraction jumps from $0$ to $0.68$ as slack increases,
never passing through the two-branch depth ($0.125$, dashed). Two binary
decisions quantize feasibility; reaching $0.125$ needs a four-option
second decision (four-plus orders), which distributes the decisive action.
The magnitude cannot transfer without hand-carving the two-branch
structure.}
\label{fig:2order}
\end{figure}

The closest scheduling analogue is two orders with a tight deadline: one
decisive early action (who runs first) and a cliff (the lag-critical
order misses if it is not first). There the correction \emph{does} help,
directionally and consistently (Figure~\ref{fig:2order}A: paired
$\Delta$AUC $+0.004$, $5/0$ non-tied seeds, $p = 0.03$, never harmful) ---
the structure and sign transfer. But the \emph{magnitude} does not, and a
second experiment shows why it cannot: adding a second binding condition
(both orders on tight deadlines) to deepen the trap makes the feasible fraction jump from $0$ to $0.68$ as the deadline loosens
(Figure~\ref{fig:2order}B), with no $\approx\!0.125$ regime in between.
Two binary decisions quantize feasibility to $\{0, \tfrac12, 1\}$; the
two-branch depth $\tfrac12 \times \tfrac14$ requires a second decision
with \emph{four} options, i.e.\ four or more orders --- at which point the
decisive structure distributes across a contention-mediated permutation
and no single early action carries the credit. Scheduling thus forces a
choice: two orders give the single decisive action but only shallow,
quantized depth; four-plus orders give the depth but distribute the
action. The single-decisive-action structure and the trap depth cannot be had together without constructing the reward by hand, as the two-branch reward is. A held correction's boundary is therefore that it pays only where a
reward isolates a single decisive early commitment, sharp enough to trap
yet dense enough to escape, whose liveness is unobservable --- a configuration realistic scheduling objectives do not admit.

\subsection{A drifting transformer learner does not rescue it}
\label{app:sched-transformer}

\begin{figure}[t]
\centering
\includegraphics[width=\linewidth]{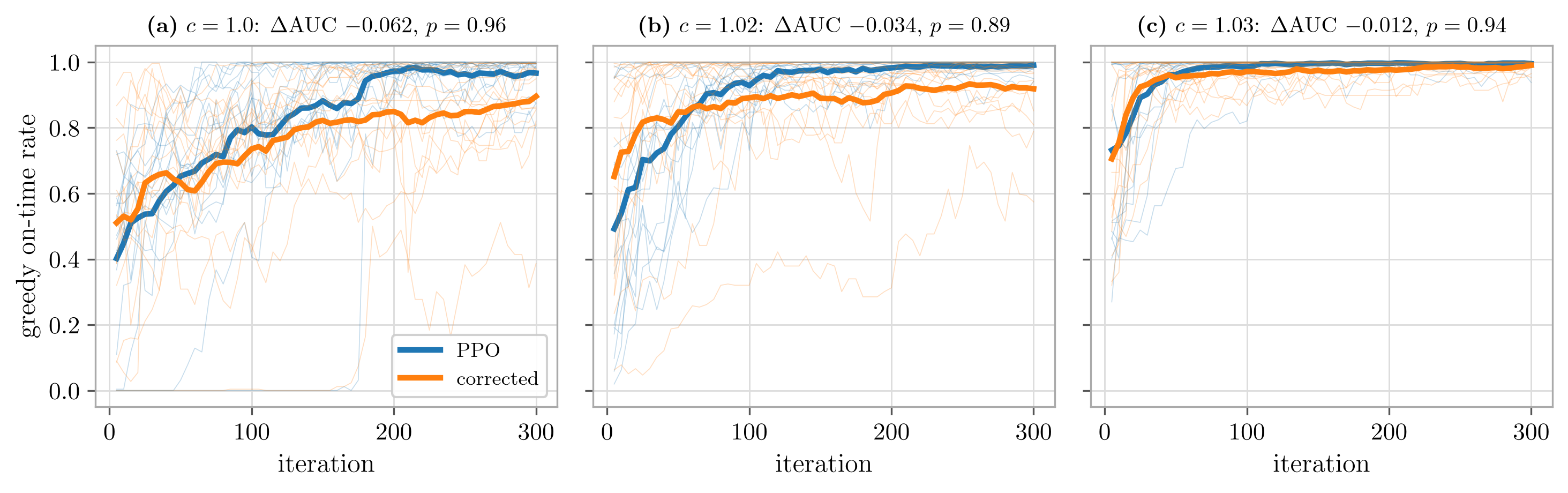}
\caption{\textbf{A drifting transformer does not rescue the correction on a
hard scheduling task.} Three-machine bottleneck flow shop, transformer
pointer policy, N-SGD, GRPO, under the scarce tight-deadline
reward at three tightnesses (on-time floor $0.14/0.35/0.56$). Greedy
on-time rate vs.\ iteration; $16$ paired seeds thin, mean bold; PPO blue,
corrected orange. At every setting PPO climbs from the floor and the corrected curve sits just below it; values in Table~\ref{tab:sched-null}.}
\label{fig:bneck}
\end{figure}

\paragraph{Transformer pointer policy.} The scheduling
nulls above all used the HGNN dispatching policy, whose pooled graph
representation barely moves per update (measured $\sigma^2 T \approx 3\times10^{-5}$ at published reuse) --- which leaves open whether they reflect a low-drift learner rather than the domain. We replaced the graph network with a \emph{transformer} pointer policy
(self-attention encoder, autoregressive attention decoder) on an NP-hard multi-machine task --- a three-machine permutation flow shop
(F3$\|C_{\max}$) with a bottleneck middle machine, $12$ jobs, varying
instances, N-SGD, and a proper GRPO group baseline ($G$ rollouts
of the same instance). The transformer drifts: $\sigma^2 T =
1.38$ at $K=12$, more than four orders of magnitude above the HGNN. The correction still does not win. Under a dense makespan reward it is a wash; under the \emph{scarce} tight-deadline reward, the only regime where a held correction has helped here (Appendix~\ref{app:sched-driftdial}), swept across three on-time floors, it consistently \emph{loses} by a small margin (Table~\ref{tab:sched-null}; Figure~\ref{fig:bneck}). At every setting PPO \emph{climbs} smoothly from the on-time floor to $\approx 0.97$ and the corrected curve tracks just below it, noisier, paying the $\alpha T$ variance premium with no stale-state bias to buy back: a competent transformer learns to schedule well, and the ``scarce'' reward is scarce only at initialization.

A held correction helps only when PPO is stuck in an absorbing trap that pins the reward near zero, and nulls or loses whenever PPO can climb; the two-branch/FATAL construction produces such a trap, and realistic scheduling objectives, with continuous progress signal, do not. The drift number makes the same point quantitatively: $\sigma^2 T = 1.38$ is real drift but an order of magnitude below the $\sigma^2 T \approx 15$--$34$ at which the drift-dial crossover fires, and reaching that band on the transformer requires raising the learning rate and loosening the clip until the update is near-divergent. Dialing the update strength up through three settings (D1: $\text{lr}\,0.25$, clip $0.5$, $K=24$; D2: $0.35/0.6/30$; D3: $0.45/0.8/40$; $16$ paired seeds each), the correction's $\Delta$AUC moves \emph{monotonically} with drift and crosses zero at D3, exactly where PPO can no longer climb (Table~\ref{tab:sched-null}). The crossover is directional, not significant, but it reproduces the HGNN's drift-dial crossover on a different architecture: the threshold is a property of the training regime, not of the learner.

\subsection{The win in detail: warmup, reuse, horizon, and dose}
\label{app:sched-win}

\begin{figure}[t]
\centering
\includegraphics[width=\linewidth]{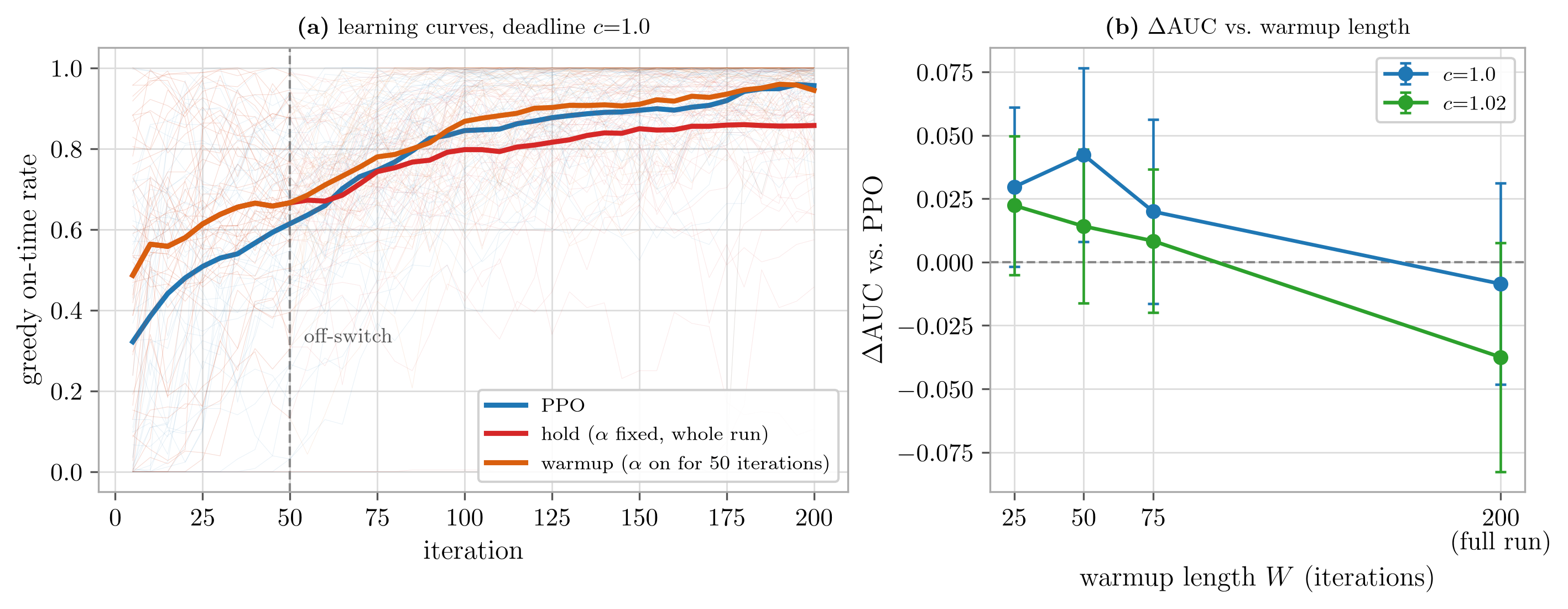}
\caption{\textbf{The correction as a warmup beats PPO; held the whole run it
loses.} Bottleneck transformer, $48$ seeds per deadline. \textbf{(A)} Greedy
on-time rate vs.\ iteration at $c=1.0$: the warmup ($\alpha$ on for $50$
iterations then off, orange) climbs above PPO (blue) early and stays $\ge$ it,
while holding a fixed $\alpha$ the whole run (red) sags below PPO late. Dotted
line marks the off-switch. \textbf{(B)} $\Delta$AUC vs.\ warmup length $W$ for
both deadlines, with hold plotted at $W=200$ (the full run); bars are $2$ SE and
the dashed line is the PPO baseline. Short warmups sit above the baseline and
holding the correction drops below it; the pooled statistics are in
Table~\ref{tab:sched-warmup}.}
\label{fig:warm}
\end{figure}

\begin{figure}[t]
\centering
\includegraphics[width=\linewidth]{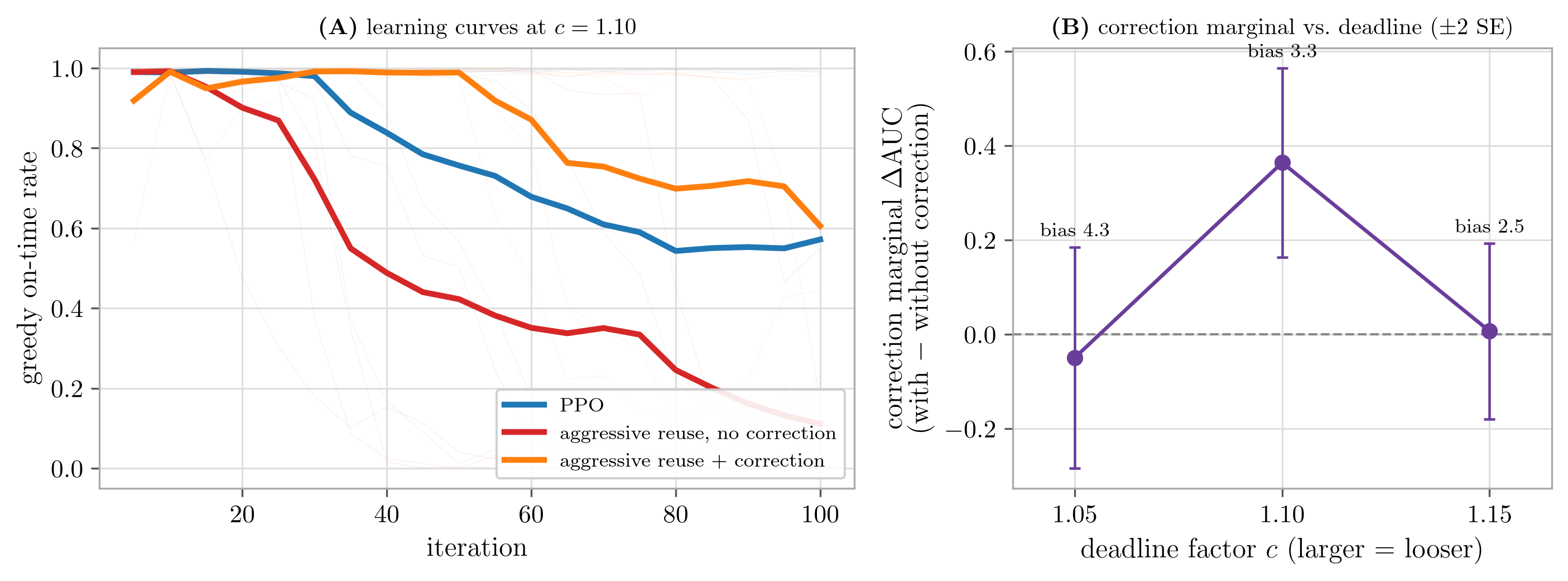}
\caption{\textbf{Disjunctive JSSP: the largest correction win, in the
longest-range credit-assignment task.} Classical JSSP ($8$ jobs, $5$ machines,
own routes, one shared bottleneck, $T = 40$, sparse binary deadline reward).
\textbf{(A)} At $c = 1.10$, aggressive reuse with the correction (orange) leads
plain PPO (blue) and reuse without the correction (red) throughout.
\textbf{(B)} Correction marginal $\Delta$AUC over reuse only vs.\
deadline ($\pm 2$ SE; early measured bias annotated): sharp at the feasible
$c = 1.10$ and vanishing on either side --- collapse at $c = 1.05$, too-loose
null at $c = 1.15$. Values are in Table~\ref{tab:sched-win}.}
\label{fig:jssp}
\end{figure}

\begin{figure}[t]
\centering
\includegraphics[width=\linewidth]{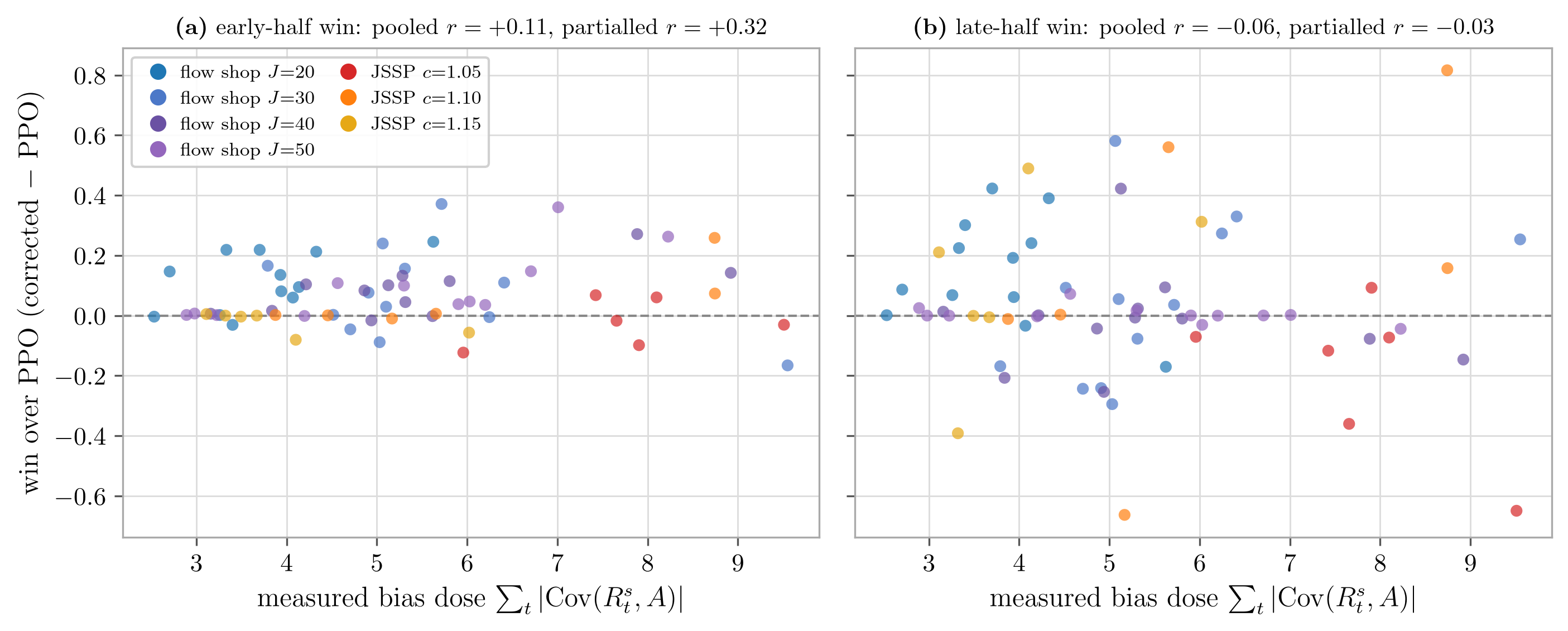}
\caption{\textbf{The early win tracks the bias dose; the late win does not.} Per configuration--seed point
($n = 66$) correction win over plain PPO vs.\ the measured prefix bias
$\sum_t |b_t|$ that aggressive reuse incurs without
correction, colored by configuration. \textbf{(Left)} First-half win rises with the
dose. \textbf{(Right)} Second-half win is uncorrelated with it --- consistent with
the mechanism being early prefix-bias cancellation that decouples once the
warmup correction switches off. Pooled and difficulty-partialled $r$ are in the
panel titles; per-configuration $r$ in Table~\ref{tab:sched-horizon}.}
\label{fig:wincorr}
\end{figure}

\begin{table}[h]
\centering
\footnotesize
\setlength{\tabcolsep}{3pt}
\begin{tabular}{lccccccccccc}
\toprule
 & & PPO & \multicolumn{3}{c}{Corrected vs.\ PPO} & \multicolumn{3}{c}{Reuse only vs.\ PPO} & \multicolumn{3}{c}{\shortstack{Correction's marginal:\\ corrected $-$ reuse only}} \\
\cmidrule(lr){4-6}\cmidrule(lr){7-9}\cmidrule(lr){10-12}
Task / deadline & $n$ & final & $\Delta$AUC & seeds & $p$ & $\Delta$AUC & seeds & $p$ & $\Delta$AUC & seeds & $p$ \\
\midrule
\multicolumn{12}{l}{\emph{Single-bottleneck flow shop, $J = 12$ (easy); $K = 30$ for the first $W = 50$ iterations, then standard $K$}} \\
$c = 1.02$ & $16$ & $0.98$ & $\boldsymbol{+0.087}$ & $\boldsymbol{14/16}$ & $\boldsymbol{0.002}$ & $\boldsymbol{+0.064}$ & $\boldsymbol{14/16}$ & $\boldsymbol{0.002}$ & $+0.023$ & $10/16$ & $0.227$ \\
\midrule
\multicolumn{12}{l}{\emph{Seven-machine flow shop, $J = 20$, three bottlenecks (hard); same $K = 30$ warmup}} \\
$c = 1.06$ & $24$ & $0.99$ & $\boldsymbol{+0.009}$ & $\boldsymbol{19/23}$ & $\boldsymbol{0.001}$ & $-0.009$ & $12/23$ & $0.500$ & $\boldsymbol{+0.018}$ & $\boldsymbol{15/21}$ & $\boldsymbol{0.039}$ \\
$c = 1.05$ & $48$ & $0.89$ & $\boldsymbol{+0.034}$ & $\boldsymbol{41/48}$ & $\boldsymbol{<10^{-4}}$ & $-0.005$ & $27/48$ & $0.235$ & $\boldsymbol{+0.039}$ & $\boldsymbol{38/48}$ & $\boldsymbol{<10^{-4}}$ \\
$c = 1.04$ & $24$ & $0.79$ & $\boldsymbol{+0.114}$ & $\boldsymbol{22/24}$ & $\boldsymbol{<10^{-4}}$ & $\boldsymbol{+0.026}$ & $\boldsymbol{17/24}$ & $\boldsymbol{0.032}$ & $\boldsymbol{+0.087}$ & $\boldsymbol{20/24}$ & $\boldsymbol{0.0008}$ \\
\midrule
\multicolumn{12}{l}{\emph{Disjunctive job shop, $8$ jobs, $5$ machines, $T = 40$; same $K = 30$ warmup}} \\
$c = 1.15$ & $6$ & $0.53$ & $+0.040$ & $3/6$ & $0.656$ & $+0.033$ & $2/5$ & $0.812$ & $+0.006$ & $2/6$ & $0.891$ \\
$c = 1.10$ & $6$ & $0.57$ & $+0.100$ & $4/6$ & $0.344$ & $-0.264$ & $0/6$ & $1.000$ & $\boldsymbol{+0.364}$ & $\boldsymbol{6/6}$ & $\boldsymbol{0.016}$ \\
$c = 1.05$ & $6$ & $0.41$ & $-0.110$ & $0/6$ & $1.000$ & $-0.060$ & $3/6$ & $0.656$ & $-0.050$ & $3/6$ & $0.656$ \\
\bottomrule
\end{tabular}
\caption{The aggressive-reuse warmup recipe: $K = 30$ for the first $50$ iterations, then plain PPO at the standard $K$, with the correction (\emph{corrected}) or without it (\emph{reuse only}). $\Delta$AUC is the paired, per-seed difference in the area under the greedy validation on-time curve; seeds counts the seeds favoring the first arm over the non-tied paired seeds, and $p$ is the one-sided exact sign test; \textbf{bold} marks a win, a positive difference with $p < 0.05$. The two arms do the same optimizer work, so the correction's marginal is its contribution beyond the reuse itself. PPO final is plain PPO's final on-time rate, which sets the headroom.}
\label{tab:sched-win}
\end{table}

\begin{table}[h]
\centering
\small
\begin{tabular}{lccccc}
\toprule
 & & PPO & \multicolumn{3}{c}{Corrected vs.\ PPO} \\
\cmidrule(lr){4-6}
Warmup length & $n$ & final & $\Delta$AUC & seeds & $p$ \\
\midrule
$W = 25$ & $96$ & $0.97$ & $\boldsymbol{+0.026}$ & $\boldsymbol{59/95}$ & $\boldsymbol{0.012}$ \\
$W = 50$ & $96$ & $0.97$ & $\boldsymbol{+0.028}$ & $\boldsymbol{58/94}$ & $\boldsymbol{0.015}$ \\
$W = 75$ & $96$ & $0.97$ & $+0.014$ & $52/96$ & $0.238$ \\
held, whole run & $96$ & $0.97$ & $-0.023$ & $42/96$ & $0.908$ \\
\bottomrule
\end{tabular}
\caption{Warmup length at the standard reuse count: bottleneck flow shop, $J = 12$, transformer policy, the correction on for the first $W$ iterations and then plain PPO (``held'': on for the whole run); deadlines $c \in \{1.0, 1.02\}$ pooled, $48$ seeds each. Columns and bolding as in Table~\ref{tab:sched-win}. There is no extra reuse here, so corrected vs.\ PPO already isolates the correction.}
\label{tab:sched-warmup}
\end{table}

\paragraph{Warmup schedule.} $\alpha = \alpha_0$ for the first $W$ iterations, then $\alpha = 0$ (plain PPO); bottleneck transformer, $48$ seeds per deadline, $c \in \{1.0, 1.02\}$, pooled in Table~\ref{tab:sched-warmup}. Per deadline, the $W = 25$ and $W = 50$ warmups remain significant at $c = 1.02$ alone ($p = 0.030$ and $0.039$). The second-half $\Delta$AUC is $-0.06$ to $-0.09$ under a held $\alpha$ and within $0.02$ of zero once $\alpha$ switches off (Figure~\ref{fig:warm}).

\paragraph{Aggressive reuse during the warmup, with a reuse-only arm.} $K = 30$ for the first $W = 50$ iterations, then the standard $K$; the reuse-only arm runs the same schedule without the correction. On the easy task (single-bottleneck flow shop, $16$ seeds) the correction at the standard $K$ gives $\Delta$AUC $+0.024$ over PPO, against the $K = 30$ values in Table~\ref{tab:sched-win}. The hard task's three bottlenecks sit at machines $\{1, 3, 5\}$.

\begin{figure}[t]
\centering
\includegraphics[width=\linewidth]{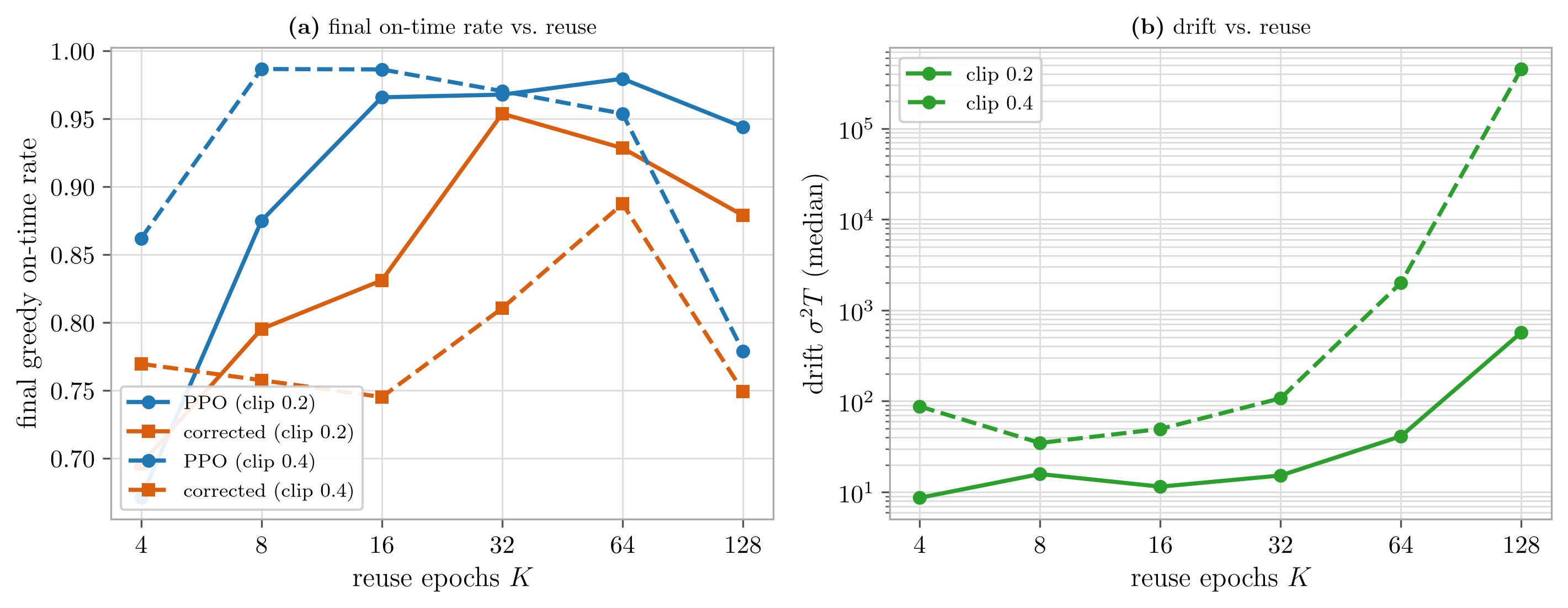}
\caption{\textbf{$K$-tolerance frontier on the easy task: reuse opens no window
for the correction.} Reuse $K$ held the whole run, swept $4$--$128$, at
$\text{clip}=0.2$ (solid) and $0.4$ (dashed). \textbf{(A)} Final greedy on-time
vs.\ $K$: PPO (blue) is robust --- it peaks at $K=64$ and barely drops by
$K=128$ at $\text{clip}=0.2$, failing only at $K=128$/$\text{clip}=0.4$ --- and
the corrected update (orange) sits at or below PPO at nearly every $K$.
\textbf{(B)} Drift $\sigma^2 T$ (median, log scale) climbs steeply with $K$ and
explodes to $\sim\!10^6$ at $K=128$/$\text{clip}=0.4$: the variance regime that
sinks the correction. Clip already contains the reuse bias on an easy task, so
the correction only adds variance --- its value needs hard credit assignment
(Figure~\ref{fig:main}c), not merely high $K$.}
\label{fig:ktol}
\end{figure}

\paragraph{$K$-tolerance frontier (easy task).} $K$ held for the whole run, swept $4$--$128$ at clip $0.2$ and $0.4$, with the drift $\sigma^2 T$ measured alongside (Figure~\ref{fig:ktol}). PPO is robust to reuse at clip $0.2$ and fails only at $K = 128$ under clip $0.4$; the corrected update sits at or below PPO at every $K$ but one, while its drift climbs by five orders of magnitude across the sweep.

\begin{figure}[t]
\centering
\includegraphics[width=\linewidth]{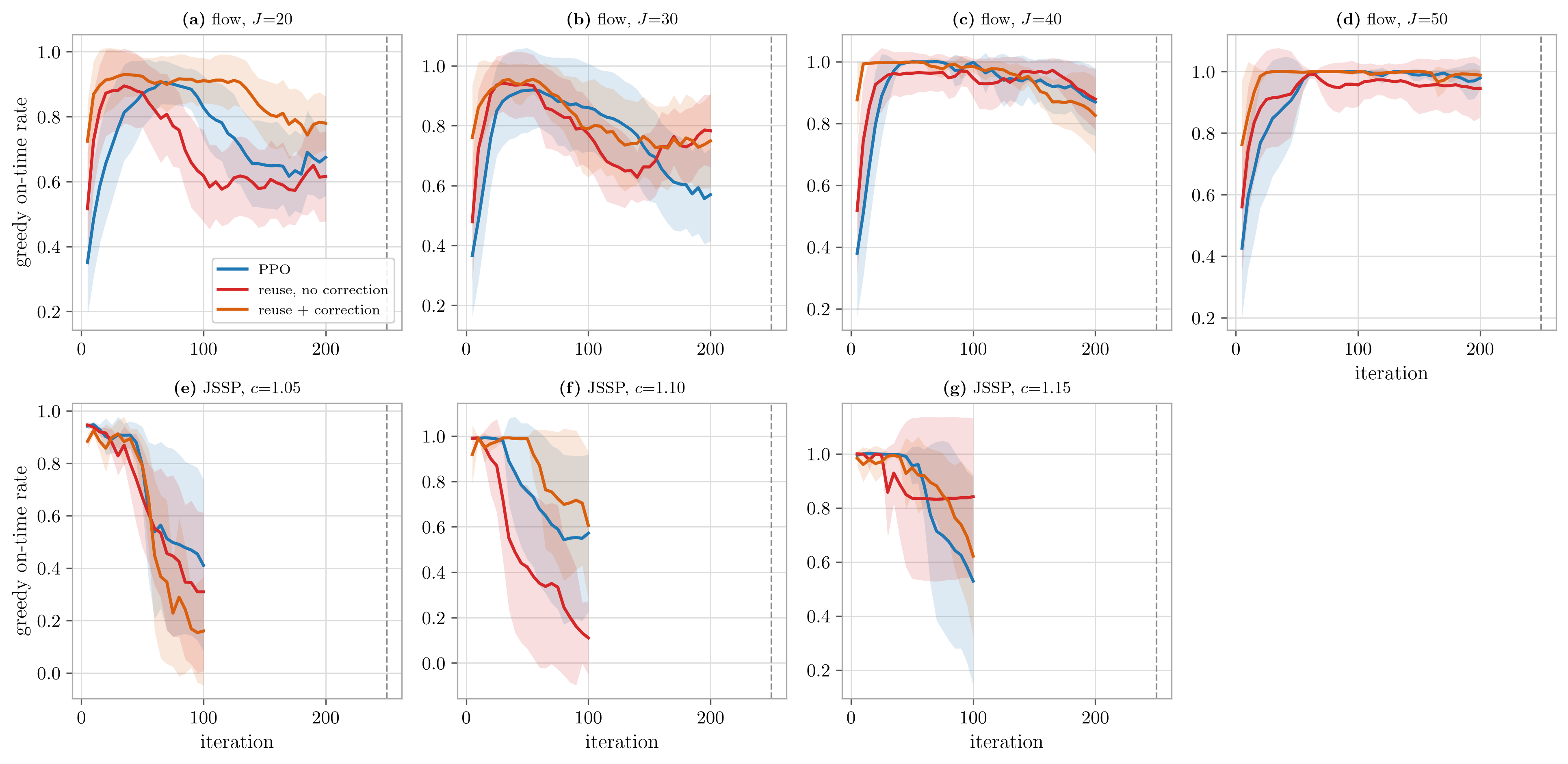}
\caption{\textbf{Faster early, same endpoint --- the sample-efficiency win
across horizon and a second task family.} Learning curves (mean $\pm 2$ SE) for
plain PPO (blue), aggressive reuse without the correction (red), and aggressive
reuse with the correction (orange); dotted line marks the end of the correction
warmup. In the long-horizon flow shop ($J = 40, 50$) the correction leads
through the warmup window and PPO converges to the same plateau afterward
(sample efficiency, not a better policy). The durable exception is disjunctive
JSSP at $c = 1.10$, where the correction stays ahead throughout.}
\label{fig:curves}
\end{figure}

\begin{figure}[t]
\centering
\includegraphics[width=\linewidth]{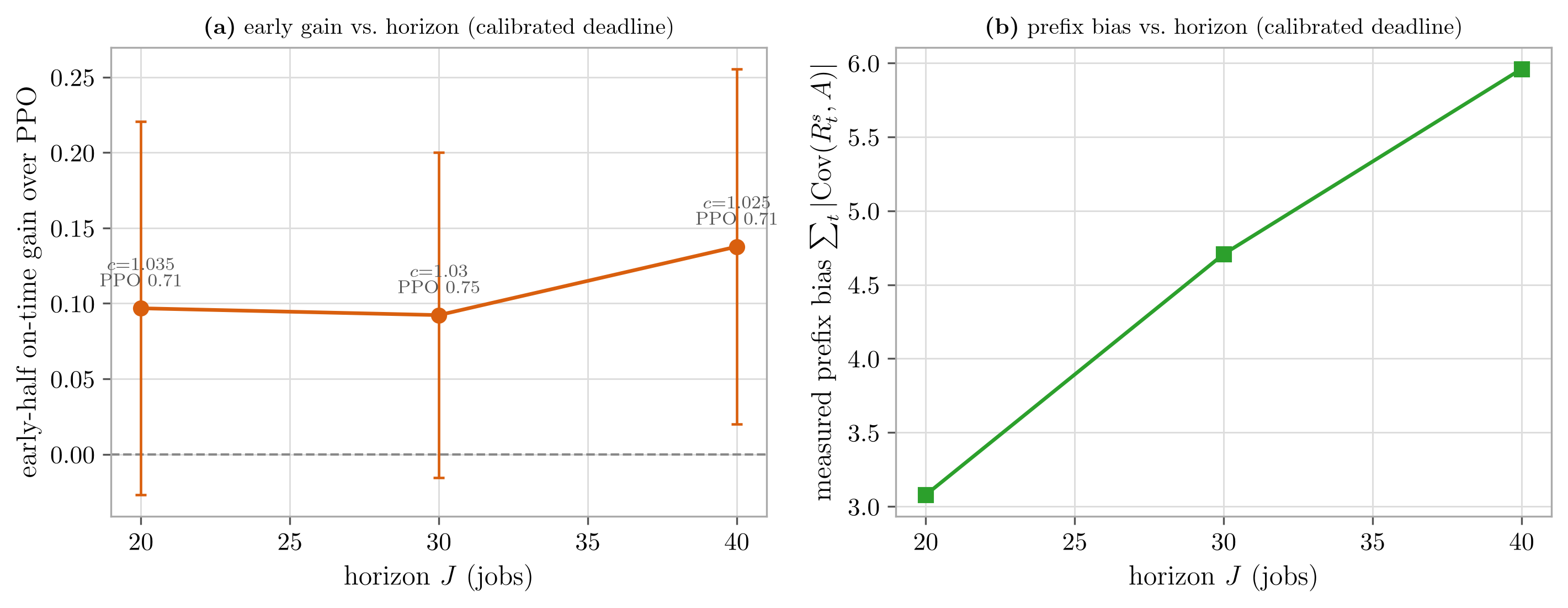}
\caption{\textbf{Horizon drives the bias dose; the win is headroom-gated.}
Ceiling-controlled horizon sweep: for each $J$ the deadline $c$ is calibrated
(PPO-only, blind to the corrected arm) so PPO's final on-time sits in a mid-band
($J = 50$ excluded --- no in-band cell). \textbf{(A)} Correction early-half win
over PPO vs.\ $J$ at the held ceiling: positive and roughly flat. \textbf{(B)}
Measured prefix bias vs.\ $J$: rises monotonically even though the ceiling is
fixed --- the dose is horizon-driven, while the realised win is capped by
headroom (here held constant). Values are in Table~\ref{tab:sched-horizon}.}
\label{fig:horizon-cal}
\end{figure}

\paragraph{Horizon sweep and calibration.} Multi-bottleneck flow shop, $J \in \{20, 30, 40, 50\}$ jobs (setting $T$), $12$ seeds each at fixed $c = 1.03$ (Table~\ref{tab:sched-horizon}, first block; Figure~\ref{fig:curves}). Because PPO's final on-time rises with $J$ at fixed $c$, the sweep partly conflates trajectory length with PPO's AUC headroom. A PPO-only calibration ($8$ seeds, blind to the corrected arm, commit \texttt{3a3c0a1}) therefore chooses, for each $J$, the deadline in a grid whose PPO final on-time is closest to $0.65$, accepting it only inside $[0.50, 0.85]$; $J = 50$ is excluded because its feasibility jumps from $0.25$ at $c = 1.01$ to $\approx 1.0$ at $c = 1.03$ with no grid point in the band (Table~\ref{tab:sched-horizon}, second block; Figure~\ref{fig:horizon-cal}).

\paragraph{Disjunctive job shop.} Machine routes are drawn at random per job, $T = J \cdot M = 40$, the same binary deadline reward, $6$ seeds (Table~\ref{tab:sched-win}, last block; Table~\ref{tab:sched-horizon}, third block; Figure~\ref{fig:jssp}). At $c = 1.05$ PPO itself falls from $0.94$ to $0.41$ over training, with no within-group signal for GRPO. $16$ seeds at $c = 1.10$ would be needed before the magnitude is headlined.

\begin{figure}[t]
\centering
\includegraphics[width=\linewidth]{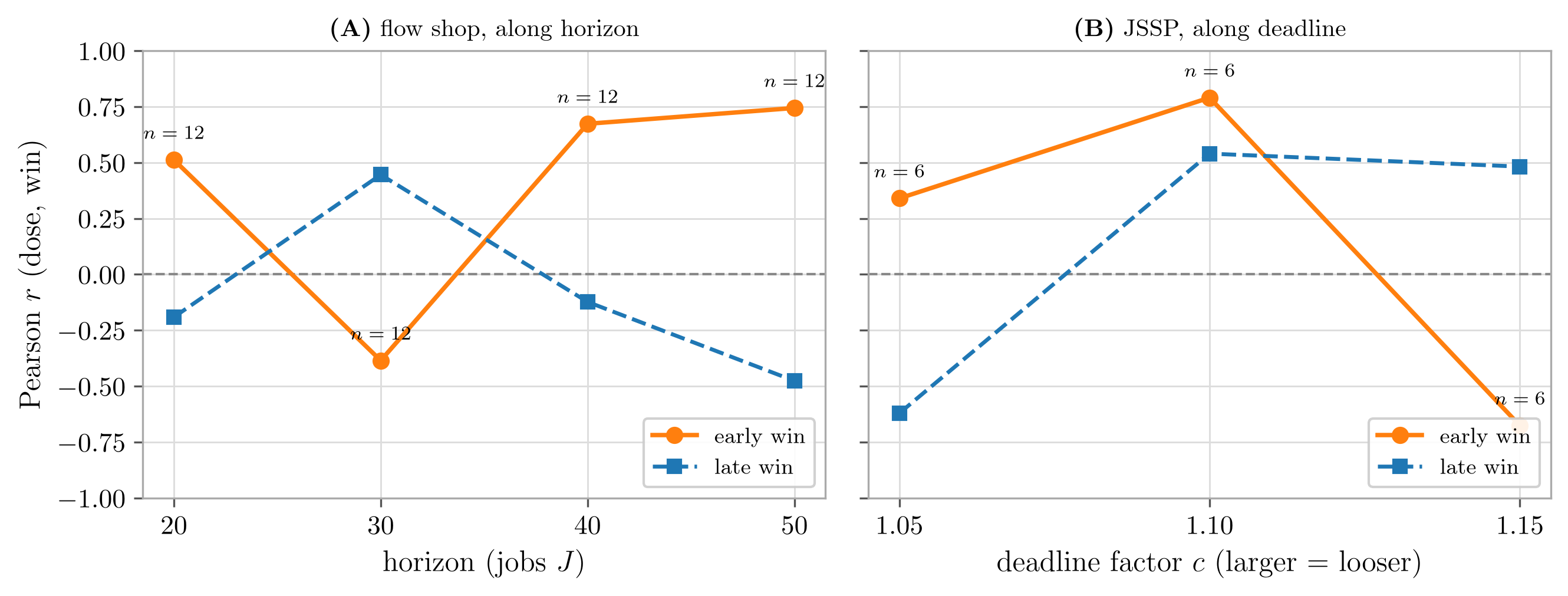}
\caption{\textbf{The early-win--dose coupling peaks where the correction wins.}
Per-config within-seed Pearson $r$ between the measured bias dose and the
correction's win over plain PPO, early (orange, solid) vs.\ late (blue, dashed),
plotted along each task's difficulty axis. \textbf{(A)} Flow shop: the early
correlation is highest at long horizon $J$. \textbf{(B)} JSSP: it peaks at the
feasible deadline $c = 1.10$ --- the winning cell --- and falls off on either
side. Per-config $n = 6$--$12$, so the late correlations scatter widely around
zero; the axis-wise pattern of the early correlation, not any single point, is
the message. Values are in Table~\ref{tab:sched-horizon}.}
\label{fig:wincorr-bycfg}
\end{figure}

\paragraph{Dose regression.} The dose of Section~\ref{sec:sched-win} is averaged over the warmup window per seed. Difficulty is partialled out by $z$-scoring dose and win within configuration before pooling the $n = 66$ points; the early-half correlation is positive in $5$ of $7$ configurations. The per-configuration correlations (Table~\ref{tab:sched-horizon}; Figure~\ref{fig:wincorr-bycfg}) are point estimates on $6$--$12$ seeds and scatter widely; the early correlation is largest on the job-shop winning cell. A correlation does not by itself establish the dose as the cause, and the across-task dose--response without difficulty control is flat because PPO's headroom caps the realised win.

\paragraph{Replication across environments.} The effect reproduces in sign and magnitude across a local $8$-seed and a cloud $16$- and $48$-seed environment with different PyTorch builds.

\begin{table}[t]
\centering
\small
\setlength{\tabcolsep}{4.5pt}
\begin{tabular}{lcccccccc}
\toprule
 & & PPO & \multicolumn{2}{c}{Lead over PPO} & Marginal & & \multicolumn{2}{c}{$r$(dose, win)} \\
\cmidrule(lr){4-5}\cmidrule(lr){8-9}
Configuration & $n$ & final & first half & second half & $\Delta$AUC & Dose & early & late \\
\midrule
\multicolumn{9}{l}{\emph{Multi-bottleneck flow shop, horizon sweep at fixed $c = 1.03$}} \\
$J = 20$ & $12$ & $0.67$ & $+0.115$ & $+0.149$ & $+0.178$ & $3.75$ & $+0.51$ & $-0.19$ \\
$J = 30$ & $12$ & $0.57$ & $+0.071$ & $+0.050$ & $+0.045$ & $5.53$ & $-0.39$ & $+0.45$ \\
$J = 40$ & $12$ & $0.87$ & $+0.083$ & $-0.016$ & $+0.021$ & $5.41$ & $+0.67$ & $-0.12$ \\
$J = 50$ & $12$ & $0.98$ & $+0.092$ & $+0.003$ & $+0.049$ & $5.27$ & $+0.74$ & $-0.48$ \\
\midrule
\multicolumn{9}{l}{\emph{Same task, deadline calibrated per $J$ to hold PPO's ceiling fixed}} \\
$J = 20$, $c = 1.035$ & $8$ & $0.71$ & $+0.097$ & $+0.077$ & $+0.109$ & $3.08$ & $+0.03$ & $-0.38$ \\
$J = 30$, $c = 1.03$ & $8$ & $0.75$ & $+0.092$ & $+0.055$ & $+0.192$ & $4.71$ & $+0.07$ & $+0.09$ \\
$J = 40$, $c = 1.025$ & $8$ & $0.71$ & $+0.138$ & $-0.092$ & $+0.158$ & $5.96$ & $-0.19$ & $-0.55$ \\
\midrule
\multicolumn{9}{l}{\emph{Disjunctive job shop, $T = 40$}} \\
$c = 1.05$ & $6$ & $0.41$ & $-0.023$ & $-0.196$ & $-0.050$ & $7.76$ & $+0.34$ & $-0.62$ \\
$c = 1.10$ & $6$ & $0.57$ & $+0.055$ & $+0.144$ & $+0.364$ & $6.11$ & $+0.79$ & $+0.54$ \\
$c = 1.15$ & $6$ & $0.53$ & $-0.023$ & $+0.102$ & $+0.006$ & $3.95$ & $-0.68$ & $+0.48$ \\
\bottomrule
\end{tabular}
\caption{Horizon and dose detail behind Section~\ref{sec:sched-win}. Lead over PPO is the warmup-corrected run's mean greedy on-time minus plain PPO's over the first and second halves of training; marginal $\Delta$AUC is over the reuse-only $K = 30$ arm; dose is the prefix bias $\sum_t |b_t|$ on that arm, averaged over the warmup window; $r$ is the within-configuration Pearson correlation across seeds between the dose and the half-wise lead. In the calibrated block the deadline is chosen per $J$ by a PPO-only calibration (see text); $J = 50$ has no deadline in the calibration band and is excluded.}
\label{tab:sched-horizon}
\end{table}


\section{Computing infrastructure and assets}
\label{app:compute}

All experiments ran on Google Cloud virtual machines and a local workstation. The transformer testbed experiments and the pretrained-LM study ran on \texttt{g2-standard-8} machines with one NVIDIA L4 GPU each; the tabular testbed, the derivation checks, and the scheduling experiments ran on CPU only, on multi-core cloud machines (\texttt{c2-standard} family) and locally. Scheduling seeds were sharded across machines and the result files merged; because random-number streams differ across software environments, cross-environment replications agree in sign but not exactly in magnitude (Appendix~\ref{app:sched-win}). The pretrained model is Qwen2.5-0.5B~\citep{qwen2024qwen25}, released under the Apache~2.0 license. The HGNN dispatcher builds on the public implementation of \citet{song2022flexible}, released under the Apache~2.0 license and included, with our modifications, in the code supplement. All scheduling instances are generated synthetically, with that implementation's instance generator for the flexible job shop; no external dataset is used.

\end{document}